\documentclass[a4paper,fleqn]{cas-sc}

\usepackage[authoryear,longnamesfirst]{natbib}
\usepackage{lineno}
\usepackage{amsthm}
\usepackage{enumitem}
\usepackage{microtype}
\usepackage{algorithm}
\usepackage{algpseudocode}
\usepackage{adjustbox}
\usepackage{natbib}
\usepackage{lastpage}
\usepackage[skins,breakable]{tcolorbox}
\usepackage[normalem]{ulem}
\usepackage{newunicodechar}
\newunicodechar{–}{--}
\hypersetup{hidelinks}
\ExplSyntaxOn
\bool_gset_true:N \g_stm_nologo_bool
\ExplSyntaxOff

\newif\ifshowrevisions
\showrevisionstrue
\definecolor{revisionDeleted}{HTML}{B22222}
\definecolor{revisionAdded}{HTML}{005BBB}
\definecolor{revisionComment}{HTML}{7A2490}
\DeclareRobustCommand{\revdel}[1]{\ifshowrevisions{\color{revisionDeleted}\sout{#1}}\fi}

\newcolumntype{L}[1]{>{\raggedright\arraybackslash}p{#1}}

\newtheorem{theorem}{Theorem}
\newtheorem{proposition}{Proposition}
\newtheorem{corollary}{Corollary}
\newtheorem{assumption}{Assumption}
\newtheorem{lemma}{Lemma}

\definecolor{rpiTheorem}{HTML}{BC6000}
\definecolor{rpiCorollary}{HTML}{1608B5}
\tcbset{rpi result/.style={
  enhanced,breakable,
  boxrule=0.8pt,arc=2mm,
  left=8pt,right=8pt,top=5pt,bottom=5pt,
  before skip=10pt,after skip=10pt,
  fonttitle=\bfseries,coltitle=white,
  toptitle=4pt,bottomtitle=4pt
}}
\tcolorboxenvironment{theorem}{
  rpi result,
  colframe=rpiTheorem,colback=rpiTheorem!3!white
}
\tcolorboxenvironment{proposition}{
  rpi result,
  colframe=rpiCorollary,colback=rpiCorollary!3!white
}
\tcolorboxenvironment{corollary}{
  rpi result,
  colframe=rpiCorollary,colback=rpiCorollary!3!white
}
\newcommand{\method}{\textup{PI-Init}}
\newcommand{\G}{\mathcal{G}}
\newcommand{\Lpi}{\mathcal{L}_{\mathrm{PI}}}
\newcommand{\Ah}{\widehat{A}}
\newcommand{\tr}{\operatorname{tr}}
\newcommand{\E}{\mathbb{E}}
\usepackage{xcolor}

\newcommand{\SL}[1]{\textcolor{purple}{[SL: #1]}}
\renewcommand{\SL}[1]{}
\begin{document}

\shorttitle{Physics-Informed Initialization for Neural Operators}
\shortauthors{S. Zhang et al.}

\title[mode=title]{Rethinking How to Leverage Cheap Physics for Data-Efficient Neural Operators}

\author[1]{Shuo Zhang}
\ead{shuo.zhang@ttu.edu}

\author[1]{Senwei Liang}[orcid=0000-0002-3558-6828]
\cormark[1]
\ead{senwei.liang@ttu.edu}

\affiliation[1]{organization={Department of Mathematics and Statistics, Texas Tech University},
            city={Lubbock},
            postcode={79409},
            state={TX},
            country={USA}}

\cortext[1]{Corresponding author.}

\begin{abstract}
Neural operators learn mappings from PDE inputs, such as initial conditions and coefficient fields, to the corresponding solutions. Once trained, they can rapidly predict solutions for new inputs. However, obtaining input--solution pairs for supervised training often requires expensive numerical simulations, whereas PDE inputs and governing equations provide inexpensive physical information. This raises a question: how can cheap physical information reduce the need for labeled solutions? A natural approach is to use physics to select inputs for labeling, but our acquisition experiments show no consistent advantage over random sampling. This observation motivates exploring an alternative: using cheap physics to improve neural operator initialization before supervised training. We propose Physics-Informed Initialization (PI-Init), a simple yet effective method that pretrains neural operators on unlabeled inputs using physics-informed losses and retains the learned parameters for supervised fine-tuning on the labeled dataset. A linearized analysis shows that pretraining can improve downstream prediction accuracy by reducing initialization errors that limited labeled data cannot effectively correct. Experiments across diverse PDE benchmarks and neural operator architectures demonstrate improvements in predictive accuracy, out-of-distribution generalization, and label efficiency. Representative results show error reductions of up to 62.7\% at the same label budget and comparable or better accuracy with 90\% fewer labels, relative to supervised-only training.
\end{abstract}


\begin{highlights}
\item Cheap-physics-guided selection does not consistently beat random sampling
\item Physics-Informed Initialization (PI-Init) uses cheap physics for label-free pretraining
\item Theoretical analysis explains how initialization affects prediction risk 
\item PI-Init improves accuracy, generalization, and label efficiency
\end{highlights}


\maketitle


\section{Introduction}
\label{sec:intro}

Many scientific and engineering applications require solving partial differential equations (PDEs) repeatedly for different problem instances. Within a given PDE family, each instance is specified by data or parameters, such as initial conditions, coefficient fields, or geometric configurations, which we collectively refer to as \emph{PDE inputs}. Conventional numerical methods typically require a separate solve for each new input configuration, making repeated evaluations computationally expensive. Neural operators address this challenge by learning mappings from PDE inputs to the corresponding solutions~\citep{deeponet,fno,neuraloperator}. Once trained, they can rapidly predict solutions for new inputs. Solution labels may require expensive numerical simulations, whereas PDE inputs and governing equations can provide additional training information without reference solutions. This difference in cost creates an opportunity to use readily available inputs and physical constraints to reduce dependence on expensive solution labels.

Cheap physical information can support data-efficient operator learning through both input selection and model training. For input selection, physics-based acquisition scores help identify inputs for expensive solution labeling, alongside uncertainty- and feature-based criteria used in active learning and coreset selection~\citep{musekamp2025,polanska2026,picore}. For model training, physics-informed operator learning uses governing-equation residuals and applicable initial- or boundary-condition constraints as training signals on unlabeled PDE inputs~\citep{pino,physicsdeeponet}. These two uses are compatible. For example, PICore performs a physics-informed warm-up to select a coreset and then restores the original initialization before supervised training~\citep{picore}. Physics-pretrained parameters can also be retained for subsequent supervised adaptation, as demonstrated by LoRA-based supervised fine-tuning with teacher distillation in a continual-learning framework~\citep{wang2026replay}. These approaches motivate examining both the value of physics for selecting inputs to label and its role in shaping the initialization from which supervised learning begins.

This distinction raises a fundamental question: \emph{is cheap physics more valuable when used to select inputs for labeling or when incorporated directly into neural operator training?} The Burgers acquisition experiments (Section~\ref{sec:res-selection}) provide initial insight: residual-, uncertainty-, and feature-based selection strategies show no consistent advantage over random sampling under the evaluated settings, even though normalized residual scores correlate with current prediction errors. These observations suggest that inputs which are difficult for the current model to predict are not necessarily the ones whose solution labels most improve supervised learning. This motivates studying an alternative use of cheap physics: improving the operator's initialization through pretraining while acquiring solution labels for uniformly sampled inputs.

Motivated by this perspective, this paper proposes \emph{Physics-Informed Initialization} (\method{}), a two-stage training strategy that transfers physics-pretrained parameters to downstream supervised neural operator learning with limited labeled data. Specifically, \method{} first pretrains the operator on cheap, unlabeled PDE inputs using governing-equation residuals and then retains the learned parameters for supervised fine-tuning. We construct the supervised training data by sampling PDE inputs uniformly at random and computing their corresponding solutions as labels. A linearized theoretical analysis (Section~\ref{sec:theory}) characterizes how physics-informed initialization interacts with supervision, revealing its potential benefits in parameter directions weakly constrained by labeled data. A complementary mode-wise analysis further establishes conditions under which approximate physics can provide positive transfer.

To evaluate \method{}, we conduct extensive experiments across eight benchmarks spanning one-dimensional and two-dimensional PDEs as well as geometry-dependent fluid dynamics, using Fourier Neural Operators (FNO)~\citep{fno}, Wavelet Neural Operators~\citep{wno} (WNO), and Laplace Neural Operators (LNO)~\citep{lno}. The results demonstrate improvements in three key aspects. In \emph{accuracy}, \method{} reduces FNO prediction errors by 62.7\% on 1D Advection and 53.3\% on Burgers. Accuracy gains are also observed for WNO and LNO and on benchmarks using approximate physics-informed objectives. In \emph{out-of-distribution (OOD) generalization}, \method{} achieves lower mean prediction errors across all evaluated distribution shifts on Burgers and 2D Navier--Stokes. In \emph{label efficiency}, \method{} achieves lower prediction error on Burgers using only 200 labeled trajectories than supervised-only training using 2,000, corresponding to 90\% fewer labels. These gains are achieved without increasing inference complexity.

The main contributions are summarized as follows:
\begin{itemize}[leftmargin=*,nosep]
\item We propose \method{}, a simple yet effective two-stage strategy that combines cheap physical information with limited labeled data for data-efficient neural operator learning, without modifying the operator architecture.
\item We develop a linearized prediction-risk analysis that characterizes initialization errors remaining after supervised training and identifies conditions under which physics-informed pretraining reduces prediction risk or introduces negative transfer through physics mismatch.
\item We empirically examine the distinction between standalone physics-informed prediction accuracy and downstream initialization quality through loss-component ablations. Experiments across diverse PDE benchmarks and neural operator architectures further demonstrate improvements in predictive accuracy, OOD generalization, and label efficiency.
\end{itemize}

The remainder of this paper is organized as follows. Section~\ref{sec:related} reviews related work, Section~\ref{sec:method} introduces \method{} and its theoretical analysis, Section~\ref{sec:experiments-main} presents the experimental results, and Section~\ref{sec:conclusion} concludes the paper. Additional proofs, experimental results, and implementation details are provided in the appendices.

\section{Related Work}
\label{sec:related}

\paragraph{Neural operators and physics-informed learning}
Neural operators learn mappings between function spaces~\citep{neuraloperator}: DeepONet uses branch and trunk networks~\citep{deeponet}, and FNO parameterizes integral kernels in Fourier space~\citep{fno}. PINNs enforce governing equations on neural solution representations~\citep{pinn}, physics-informed DeepONets learn solution operators from residuals and initial or boundary conditions~\citep{physicsdeeponet}, and PINO combines data and physics losses, admits unlabeled PDE inputs, and refines predictions instance-wise~\citep{pino}. Recent work studies how to train such operators jointly with data, including loss balancing, second-order optimizers, and gradient conflicts~\citep{chen2026pinotraining}, preconditioned residual losses~\citep{wen2026preconditioned}, and physics-informed Laplace operators with virtual inputs~\citep{kim2026pilno}. These methods use residuals as training signals throughout learning and must balance them against the data loss; \method{} confines the residual to a label-free stage whose parameters initialize supervised training, separating the two objectives in time.

\paragraph{Pretraining for data-efficient operators}
Operator pretraining transfers information learned before supervised training~\cite{liang2024effective}. \citet{zhoupretrain} compare pretraining objectives, architectures, and source PDEs; \citet{chenpretrain} pretrain on unlabeled PDE inputs with masked reconstruction and super-resolution and adds similarity-based in-context prediction; and geometry-based pretraining learns occupancy embeddings from unlabeled shapes~\citep{geometrypretrain}. Label-free physics pretraining has also been used: PFEM pretrains a Transolver-based physics-informed operator solely on governing equations and uses its prediction to warm-start a finite-element solver, keeping the network weights fixed~\citep{wang2026pfem}, and a follow-up adapts such an operator with LoRA-based supervised fine-tuning and teacher distillation on error-ranked labels for continual learning~\citep{wang2026replay}. \method{} does not claim physics pretraining as new. It asks whether, under a fixed labeling budget, the information in a physics-pretrained model should be used to select inputs for labeling or retained in the parameters for supervised learning. To examine the value of retention, it samples PDE inputs uniformly at random and obtains their corresponding solutions as labels, compares physics-pretrained and random initializations on the resulting training data using supervised-only fine-tuning, and characterizes when retention helps or hurts through a fidelity boundary.

\paragraph{Data selection and active learning}
AL4PDE benchmarks uncertainty- and feature-based acquisition for neural PDE solvers~\citep{musekamp2025}, physics-based active learning acquires inputs by the residual of the current prediction~\citep{polanska2026}, and PICore selects coresets from physics-informed losses and gradients after a warm-up, whose parameters are then discarded~\citep{picore}. \method{} reverses this use of cheap physics: it samples PDE inputs uniformly at random for labeling and retains the physics-trained parameters for supervised learning, separating the value of cheap physics for selecting inputs to label from its value for training the neural operator.

\section{Method}
\label{sec:method}

\begin{figure}[pos=htbp]
\centering
\includegraphics[width=0.9\linewidth]{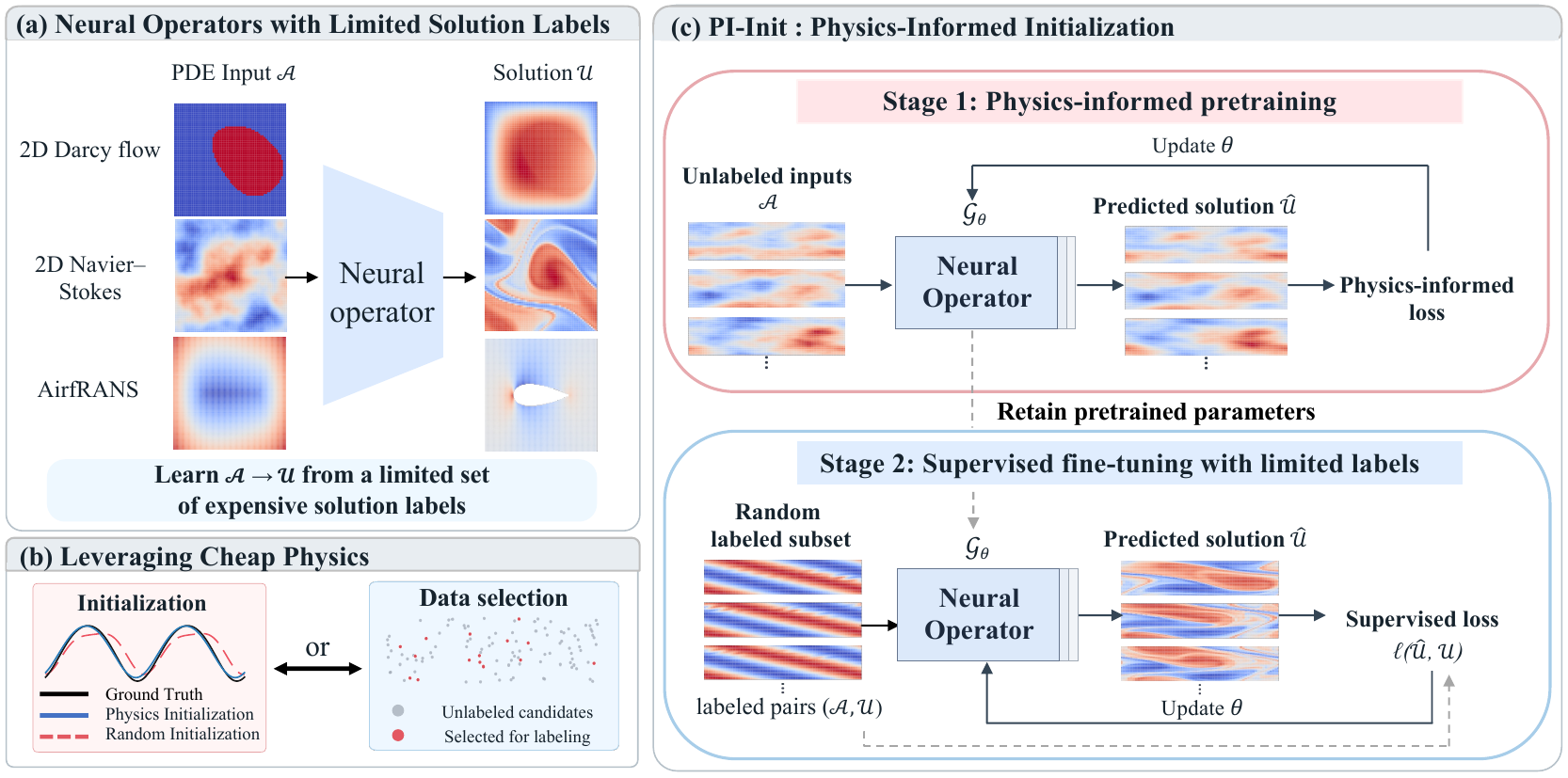}
\caption{Overview of \method{}: (a--b) motivation for
data-efficient learning and (c) the proposed two-stage framework.}
\label{fig:rpi-overview}
\end{figure}

Physics-Informed Initialization (\method{}) uses cheap physical information to learn transferable initializations for supervised neural operator training. As illustrated in Fig.~\ref{fig:rpi-overview}, neural operators typically rely on expensive input--solution pairs, while cheap physics can be exploited either for data selection or model initialization. \method{} adopts the latter strategy by pretraining on unlabeled PDE inputs using physics-informed losses and retaining the learned parameters for supervised fine-tuning. The supervised training data are constructed by sampling PDE inputs uniformly at random and computing their corresponding solutions as labels.

\subsection{Problem Setup and Model Comparisons}
\label{sec:prelim}

Operator learning aims to approximate a solution operator $\G:\mathcal A\to\mathcal X$ that maps a PDE input $a\in\mathcal A$ to its corresponding solution $u=\G(a)\in\mathcal X$. A neural operator $\G_\theta$ provides a parameterized approximation to $\G$, with trainable parameters $\theta$. Given a candidate input pool $\mathcal U=\{a_i\}_{i=1}^{M}$ and a labeling budget $B$, we consider three models that differ in how inputs are selected for labeling or how the operator is initialized:
\begin{enumerate}[leftmargin=*,nosep]
\item Supervised-only model $\G_{\theta_{\rm sup}}$.
We sample $B$ input indices uniformly at random, denote the resulting index set by $S_B$, and compute the corresponding solutions to construct
\begin{equation}
\mathcal D_B=\{(a_i,u_i):i\in S_B\}.
\end{equation}
The operator is trained on $\mathcal D_B$ from random initialization $\theta_0$ using the supervised loss. 

\item Data-selection model $\G_{\theta_{\rm select}}$.
An acquisition strategy guides the selection of inputs for labeling under the same budget $B$. Their corresponding solutions are obtained, and the operator is trained on the resulting input--solution pairs from random initialization $\theta_0$.

\item \method{} model $\G_{\theta_{\method}}$.
Physics-informed pretraining on unlabeled PDE inputs produces parameters $\theta_{\rm PI}$. Starting from $\theta_{\rm PI}$, supervised fine-tuning on the same dataset $\mathcal D_B$ used by the supervised-only model yields the final parameters $\theta_{\method}$.
\end{enumerate}

Comparing the data-selection model with the supervised-only model evaluates the effect of input selection, while comparing \method{} with the supervised-only model evaluates the effect of physics-informed initialization. The architecture, supervised objective, and downstream optimization settings are matched within each comparison.

\subsection{PI-Init: Label-Free Physics Pretraining and Supervised Fine-Tuning}
\label{sec:pi-pretraining}

Let $\mathcal U_{\rm PI}$ denote a set of $N$ unlabeled PDE inputs drawn from the training-input distribution, excluding held-out test inputs. For the prediction $\hat u_\theta(a)=\G_\theta(a)$, the physics-informed loss combines the PDE residual with applicable initial- and boundary-condition constraints:
\begin{equation}
\widehat{\Lpi}(\theta)
=
\frac{1}{N}
\sum_{a\in\mathcal U_{\rm PI}}
\left[
\lambda_{\rm PDE}
\|\mathcal F_a(\hat u_\theta(a))\|_2^2
+
\lambda_{\rm IC}
\|\mathcal I_a(\hat u_\theta(a))\|_2^2
+
\lambda_{\rm BC}
\|\mathcal B_a(\hat u_\theta(a))\|_2^2
\right],
\label{eq:pi_loss}
\end{equation}
where $\mathcal F_a$, $\mathcal I_a$, and $\mathcal B_a$ are the corresponding residual operators, with weights $\lambda_{\rm PDE}$, $\lambda_{\rm IC}$, and $\lambda_{\rm BC}$. Residuals are evaluated over predicted space-time solutions for time-dependent PDEs and spatial solutions for steady-state problems.

Starting from random initialization $\theta_0$, we optimize the physics-informed objective in Eq.~\eqref{eq:pi_loss} for $T_{\rm PI}$ epochs to obtain $\theta_{\rm PI}$. We then retain the pretrained parameters and initialize a fresh optimizer and learning-rate scheduler for supervised fine-tuning using
\begin{equation}
F_B(\theta)
=
\frac{1}{B}
\sum_{i\in S_B}
\ell(\G_\theta(a_i),u_i),
\label{eq:sup_loss}
\end{equation}
where $\ell$ denotes the loss for an input--solution pair. After $T_{\rm sup}$ epochs, the resulting parameters are denoted by $\theta_{\method}$.

The supervised-only baseline optimizes the same objective from the random initialization $\theta_0$ on the identical labeled dataset $\mathcal D_B$ for $T_{\rm sup}$ epochs. Both methods use matched supervised optimization settings and differ only in the initialization of the supervised stage. Since \method{} preserves the original operator architecture, it introduces no additional inference computation.


\subsection{Theoretical Analysis}
\label{sec:theory}

To understand how physics-informed initialization affects downstream supervised learning, the analysis first establishes a bias--variance decomposition that isolates the contribution of initialization to prediction risk. It then characterizes how physics-informed pretraining modifies this contribution through its interaction with supervised optimization. The analysis is developed under a linearized operator model, with complete proofs provided in Appendix~\ref{app:theory-proofs}.

\begin{assumption}[Linearized operator model]
\label{ass:lm}
The neural operator is represented as $\G_\theta(a)=\Phi(a)\theta$, where $\Phi(a)\in\mathbb R^{m\times d}$ denotes a fixed feature representation, as in a linearized or lazy-training regime. Labels satisfy $u=\Phi(a)\theta^\star+\xi$, with $\E[\xi\mid a]=0$ and $\operatorname{Cov}[\xi\mid a]=\omega^2(a)I_m$. Label noises are conditionally independent across samples given the inputs. Test inputs follow a distribution $\mu$, with $\Sigma=\E_\mu[\Phi(a)^\top\Phi(a)]\succ0$ and prediction risk $\mathcal E(\theta)=\|\theta-\theta^\star\|_\Sigma^2$. For a labeled subset $S$ of size $B$, define $\Ah_S=B^{-1}\sum_{i\in S}\Phi(a_i)^\top\Phi(a_i)$.
\end{assumption}

\subsubsection{Selection and Initialization}
\label{sec:m-select}

Data selection and physics-informed initialization influence supervised learning through different mechanisms. Under bounded leverage and homoscedastic noise, the least-squares risk of uniform sampling is, with high probability, within a factor $\kappa/(1-\varepsilon)$ of the optimal full-rank input-only design under the conditions established in Appendix~\ref{app:proof-random}. This limits the potential advantage of targeted selection when the leverage is sufficiently controlled. Moreover, acquisition utility measures the reduction in test-weighted posterior uncertainty  (Appendix~\ref{app:proof-difficulty}). A complementary Burgers stability estimate connects PDE residuals to prediction errors (Appendix~\ref{app:burgers-energy}), illustrating why residual-based difficulty does not necessarily translate into greater labeling value. These observations motivate examining how physics-informed pretraining improves learning from a fixed labeled subset.

\subsubsection{Initialization Bias and Physics Transfer}
\label{sec:m-where}
\label{sec:m-howmuch}

Consider a fixed labeled subset $S$ and $T$ gradient steps on the quadratic supervised objective $F_B(\theta)=\frac{1}{2B}\sum_{i\in S}\|\Phi(a_i)\theta-u_i\|_2^2$, starting from initialization $\theta_{\rm init}$ with step size $0<\eta\le1/\lambda_{\max}(\Ah_S)$. Let $\theta_T$ denote the resulting parameters and define $P_T=(I-\eta\Ah_S)^T$.

\begin{assumption}[Information boundary]
\label{ass:boundary}
Conditional on the training inputs, the initialization $\theta_{\rm init}$ and labeled subset $S$ are independent of the label noise $\{\xi_i\}_{i\in S}$. This condition holds for random initialization, physics-informed pretraining using only inputs and governing equations, and input-only data selection.
\end{assumption}

\begin{theorem}[Bias--variance decomposition under gradient descent]
\label{thm:decomp}
Under Assumptions~\ref{ass:lm}--\ref{ass:boundary}, the expected prediction risk after $T$ supervised gradient steps satisfies
\begin{equation}
\E_\xi\,\mathcal E(\theta_T)
=
\underbrace{\big\|P_T(\theta_{\rm init}-\theta^\star)\big\|_\Sigma^2}_{\text{initialization bias }\mathcal B_T(\theta_{\rm init})}
+
\underbrace{V_T(S)}_{\text{label variance}},
\label{eq:bias-variance}
\end{equation}
where $V_T(S)$ is independent of $\theta_{\rm init}$. The expectation $\E_\xi$ is taken over label noise conditional on the selected inputs, labeled subset $S$, and initialization $\theta_{\rm init}$. Furthermore, the component of $\theta_T-\theta^\star$ in $\ker\Ah_S$ remains equal to that of $\theta_{\rm init}-\theta^\star$ for every $T$.
\end{theorem}

The decomposition shows that physics-informed initialization affects downstream prediction risk through the propagated initialization bias, while the label-variance term remains unchanged. Supervised optimization contracts initialization errors along directions with positive label curvature but preserves them in the null space of $\Ah_S$. Consequently, physics-informed pretraining can have a lasting effect in parameter directions that are weakly constrained by labeled data (Appendix~\ref{app:proof-decomp}).

To quantify this effect, physics-informed pretraining is represented by a quadratic surrogate,
\begin{equation}
\widehat{\Lpi}(\theta)
=
\frac12(\theta-\tilde\theta)^\top
H_{\rm PI}(\theta-\tilde\theta)
+\mathrm{const},
\end{equation}
where $H_{\rm PI}\succeq0$ denotes the physics curvature matrix and $\tilde\theta$ is a minimizer of the surrogate objective. This formulation provides a tractable model of physics-informed optimization while allowing the physics minimizer to differ from the supervised target.

\begin{assumption}[Quadratic physics surrogate and spectral compatibility]
\label{ass:sc}
Physics-informed pretraining follows the quadratic surrogate throughout the considered optimization trajectory. The matrices $\Sigma$, $\Ah_S$, and $H_{\rm PI}$ share an orthonormal eigenbasis $\{v_k\}$, with corresponding eigenvalues $w_k$ (test weight), $s_k$ (label curvature), and $\rho_k$ (physics curvature), respectively. Starting from $\theta_0$, physics pretraining performs $T'$ gradient steps with step size $0<\eta'\le1/\max_k\rho_k$, producing $\theta_{\rm PI}$.
\end{assumption}

The shared eigenbasis assumption enables an explicit characterization of how physics pretraining and supervised learning act on the same parameter directions. Such alignment can arise in idealized translation-invariant settings, although exact compatibility of finite-sample matrices is not generally guaranteed. Define
\begin{equation}
\begin{aligned}
\alpha_k &= w_k(1-\eta s_k)^{2T}, &
c_k &= (1-\eta'\rho_k)^{T'},\\
e_k &= \langle\theta_0-\theta^\star,v_k\rangle, &
\beta_k &= \langle\tilde\theta-\theta^\star,v_k\rangle.
\end{aligned}
\end{equation}
Here, $\alpha_k$ measures the influence of initialization error remaining after supervised training, $c_k$ describes physics-induced contraction, $e_k$ denotes the initial parameter error, and $\beta_k$ captures the discrepancy between the physics minimizer and the supervised target. Let $\theta_T^{\rm Ord}$ and $\theta_T^{\method}$ denote the parameters obtained after $T$ supervised steps from $\theta_0$ and $\theta_{\rm PI}$, respectively.

\begin{theorem}[Effect of physics-informed initialization on prediction risk]
\label{thm:modewise}
Under Assumptions~\ref{ass:lm}--\ref{ass:sc}, the expected reduction in prediction risk from physics-informed initialization is
\begin{equation}
\Delta_{\rm init}
=
\E_\xi\mathcal E(\theta_T^{\rm Ord})
-
\E_\xi\mathcal E(\theta_T^{\method})
=
\sum_k\alpha_k
\left[
e_k^2-
\big(c_ke_k+(1-c_k)\beta_k\big)^2
\right].
\label{eq:modewise}
\end{equation}
For matched physics ($\beta_k=0$ for all $k$), the gain simplifies to
\begin{equation}
\Delta_{\rm init}
=
\sum_k\alpha_k(1-c_k^2)e_k^2
\ge0.
\end{equation}
For general physics, with the supervised duration $T$ fixed, the converged-pretraining gain is
\begin{equation}
\Delta_{\rm init}^{\infty}:=\lim_{T'\to\infty}\Delta_{\rm init}=\sum_{k:\rho_k>0}\alpha_k(e_k^2-\beta_k^2),
\end{equation}
so retained physics yields positive transfer at convergence if and only if
\begin{equation}
\sum_{k:\rho_k>0}\alpha_k\beta_k^2<\sum_{k:\rho_k>0}\alpha_ke_k^2.
\label{eq:fidelity-boundary}
\end{equation}
\end{theorem}

The result characterizes how physics-informed initialization interacts with the information provided by labeled data. Modes with small label curvature $s_k$ retain more of their initialization error, making them particularly sensitive to physics-induced contraction. Under matched physics, pretraining reduces these errors, whereas modes with $\rho_k=0$ remain unaffected. When the physics objective is approximate, the benefit additionally depends on the mismatch $\beta_k$, so stronger physics pretraining does not necessarily lead to better downstream performance.

The same analysis explains why standalone physics-informed prediction accuracy need not determine downstream transfer quality. The PI-only prediction error weights the pretrained errors $\epsilon_k=c_ke_k+(1-c_k)\beta_k$ by $w_k$, whereas the fine-tuned prediction risk weights them by $\alpha_k$. Since supervised optimization changes the relative importance of parameter directions, a checkpoint with lower PI-only error may not achieve lower error after fine-tuning. This distinction highlights the importance of how well physics-pretrained parameters support subsequent supervised learning, rather than how accurately the pretrained operator predicts solutions before fine-tuning (Appendix~\ref{app:proof-modewise}).

\section{Experiments}
\label{sec:experiments-main}

We numerically investigate how cheap physics can support neural operator learning when solution labels are limited. We first examine its use in input selection by comparing different acquisition strategies with random sampling (Section~\ref{sec:res-selection}). We then turn to model initialization and evaluate whether \method{} improves predictive accuracy, label efficiency, and OOD generalization. To understand the contributions of its training components and assess its computational cost, we further conduct ablation studies on physics-loss components and staged training and quantify the pretraining overhead (Sections~\ref{sec:res-rpi}--\ref{sec:res-mechanism}). Supplementary analyses supporting these evaluations and detailed experimental settings are provided in Appendices~\ref{app:additional-results} and~\ref{app:experimental_details}, respectively.

\subsection{Experimental Setup}
\label{sec:experiments}

\paragraph{Benchmarks and architectures.}
Experiments are conducted on six PDEBench tasks~\citep{pdebench}, 2D Navier--Stokes~\citep{fno}, and AirfRANS~\citep{airfrans}, covering diverse PDE systems and geometry-dependent fluid dynamics. An additional 2D Reaction--Diffusion study is provided in Appendix~\ref{app:2d-results}. Benchmark configurations and physics-informed residuals are detailed in Table~\ref{tab:benchmarks} (Appendix~\ref{app:benchmarks}). For 1D CFD and AirfRANS, physics-informed pretraining employs approximate governing equations to enable an evaluation of \method{} under approximate physical constraints. FNO~\citep{fno} serves as the primary architecture, with WNO~\citep{wno} and LNO~\citep{lno} included to assess generality across neural operator architectures.

\paragraph{Training protocol and evaluation.}
For time-dependent PDEBench tasks, FNO predicts full solution trajectories autoregressively, whereas WNO and LNO directly predict the terminal state.
In \method{}, we first perform physics-informed pretraining for 10 or 50 epochs on $N$ unlabeled inputs, denoted by PI10 and PI50, respectively.
The pretrained parameters initialize 500 epochs of supervised fine-tuning on $B$ labeled samples, with the same training settings as the randomly initialized supervised-only baseline.
For FNO, $N$ is typically equal to or a few times larger than $B$ (e.g., Burgers uses $N=50{,}000$ and $B=10{,}000$).
Predictive performance is evaluated using the mean per-sample relative $L_2$ error, computed over predicted trajectories for FNO and terminal states for WNO and LNO.
AirfRANS reports mean squared error (MSE) on normalized pressure fields.
Results are reported as mean $\pm$ standard deviation across random seeds using final-epoch models.
Detailed training configurations, including per-task values of $N$ and $B$, are provided in Appendix~\ref{app:experimental_details} (Tables~\ref{tab:benchmarks}--\ref{tab:fno-data-counts}).

\subsection{Physics-Guided Data Selection Versus Random Sampling}
\label{sec:res-selection}

To assess the value of data selection, we test whether physics-guided input selection yields lower final test error than uniform random sampling under the same labeling budget. Experiments are conducted on Burgers with fixed and variable viscosity. Each run initially obtains solutions for $P=200$ inputs, with 160 input--solution pairs used for training and 40 for validation. Labels are then acquired over eight rounds of 100 additional inputs each, giving a total labeling budget of $B=1000$. At each round, the acquisition strategies use probe FNO models trained for 50 epochs to evaluate unlabeled candidates, whereas the random baseline samples inputs uniformly. After acquisition, a new FNO is trained from random initialization for 200 epochs on the acquired training data. Using the same final training configuration and discarding the probe parameters allows the comparison to assess the value of input selection independently of parameter transfer. Each strategy is evaluated over three acquisition seeds.


The acquisition strategies include Normalized RMS PRE and Absolute MAE PRE, which combine 90 uniformly sampled trajectories with 10 score-guided selections per round~\citep{polanska2026}; PRE--SBAL, which uses mean absolute PDE residuals for score-proportional sampling~\citep{polanska2026,musekamp2025}; QbC--SBAL, which applies the same sampling rule to predictive disagreement between two independently initialized models~\citep{musekamp2025}; and LCMD, which performs greedy diversity-based selection using learned FNO features~\citep{holzmuller2023framework}. Table~\ref{tab:selection-mean} reports the test errors of the FNO models retrained after acquisition. Random sampling achieves the lowest mean test error under both viscosity settings.


To interpret these acquisition results, we further examine the relationship between residual-based scores and the probe model's current prediction errors. Across 960 pooled score--error pairs, Normalized RMS PRE and an additional relative MAE residual score achieve Spearman correlations of $0.663$ and $0.715$, respectively, with the probe prediction error (Appendix~\ref{app:selection-diagnostics}). These correlations indicate that normalized residual scores can help identify inputs on which the current model is less accurate. However, identifying such inputs does not necessarily identify those whose labels will most improve the model after retraining. This distinction between prediction difficulty and labeling utility is consistent with Proposition~\ref{prop:difficulty} (Appendix~\ref{app:proof-difficulty}). Together with the acquisition results, these diagnostics motivate using cheap physics to improve model initialization rather than relying solely on input selection.

\begin{table}[htbp]
\centering
\small
\caption{Data-selection strategies on Burgers with fixed and variable viscosity ($B=1000$). Normalized RMS PRE and Absolute MAE PRE are variants inspired by physics-based active learning~\citep{polanska2026}. Bold indicates the lowest mean error in each column.}

\label{tab:selection-mean}
\begin{adjustbox}{max width=\linewidth}
\begin{tabular}{lcc}
\toprule
Data-selection strategy & Fixed Viscosity & Variable Viscosity \\
\midrule
Random & $\mathbf{0.1484\pm0.0159}$ & $\mathbf{0.2842\pm0.0036}$ \\
Normalized RMS PRE~\citep{polanska2026} & $0.1514\pm0.0020$ & $0.2897\pm0.0046$ \\
Absolute MAE PRE~\citep{polanska2026} & $0.1610\pm0.0104$ & $0.3013\pm0.0301$ \\
PRE--SBAL~\citep{musekamp2025} & $0.1656\pm0.0185$ & $0.2902\pm0.0081$ \\
QbC--SBAL~\citep{musekamp2025} & $0.1613\pm0.0095$ & $0.3025\pm0.0040$ \\
LCMD~\citep{holzmuller2023framework} & $0.1680\pm0.0243$ & $0.3773\pm0.0937$ \\
\bottomrule
\end{tabular}
\end{adjustbox}
\end{table}

\subsection{Predictive Performance Across PDEs and Neural Operator Architectures}
\label{sec:res-rpi}

\begin{table}[pos=b]
\centering
\caption{Test performance across PDE benchmarks and neural
operator architectures (mean $\pm$ SD over three seeds).
Errors are measured by relative $L_2$, except for AirfRANS,
which reports MSE on normalized pressure fields. The labeling budgets $B$ and numbers of unlabeled pretraining inputs $N$ are reported in Table~\ref{tab:fno-data-counts}.}
\label{tab:cross_architecture_comparison}
\resizebox{\textwidth}{!}{
\begin{tabular}{lcccccc}
\toprule
\multirow{2}{*}{Dataset}
& \multicolumn{2}{c}{FNO}
& \multicolumn{2}{c}{WNO}
& \multicolumn{2}{c}{LNO} \\
\cmidrule(lr){2-3}\cmidrule(lr){4-5}\cmidrule(lr){6-7}
& Supervised-only & +\method{}
& Supervised-only & +\method{}
& Supervised-only & +\method{} \\
\midrule
Burgers & $0.0480 \pm 0.0081$ & $\mathbf{0.0224 \pm 0.0031}$ & $0.0240 \pm 0.0012$ & $\mathbf{0.0212 \pm 0.0012}$ & $0.0980 \pm 0.0037$ & $\mathbf{0.0854 \pm 0.0124}$ \\
1D Reaction--Diffusion & $0.0197 \pm 0.0029$ & $\mathbf{0.0158 \pm 0.0014}$ & $0.0020 \pm 0.0013$ & $\mathbf{0.0012 \pm 0.0001}$ & $0.0047 \pm 0.0005$ & $\mathbf{0.0040 \pm 0.0003}$ \\
1D Advection & $0.0660 \pm 0.0490$ & $\mathbf{0.0246 \pm 0.0008}$ & $0.0072 \pm 0.0002$ & $\mathbf{0.0068 \pm 0.0006}$ & $0.0052 \pm 0.0005$ & $\mathbf{0.0039 \pm 0.0004}$ \\
Diffusion--Sorption & $0.0041 \pm 0.0000$ & $\mathbf{0.0035 \pm 0.0003}$ & $0.0009 \pm 0.0000$ & $\mathbf{0.0008 \pm 0.0003}$ & $0.0006 \pm 0.0000$ & $0.0006 \pm 0.0000$ \\
1D CFD & $0.1304 \pm 0.0088$ & $\mathbf{0.0818 \pm 0.0009}$ & $0.1061 \pm 0.0003$ & $\mathbf{0.1051 \pm 0.0010}$ & $0.1125 \pm 0.0017$ & $\mathbf{0.1108 \pm 0.0014}$ \\
2D Darcy Flow & $0.1366 \pm 0.0058$ & $\mathbf{0.1233 \pm 0.0065}$ & $0.1312 \pm 0.0022$ & $\mathbf{0.1180 \pm 0.0045}$ & $0.0239 \pm 0.0014$ & $\mathbf{0.0210 \pm 0.0012}$ \\
AirfRANS & $0.0351 \pm 0.0005$ & $\mathbf{0.0317 \pm 0.0004}$ & $0.1163 \pm 0.0013$ & $\mathbf{0.1092 \pm 0.0059}$ & $0.0650 \pm 0.0012$ & $\mathbf{0.0607 \pm 0.0052}$ \\
2D Navier--Stokes & $0.1395 \pm 0.0016$ & $\mathbf{0.1253 \pm 0.0021}$ & --- & --- & --- & --- \\
\bottomrule
\end{tabular}
}
\end{table}

\begin{figure}[pos=htbp]
\centering
\includegraphics[width=0.95\textwidth]{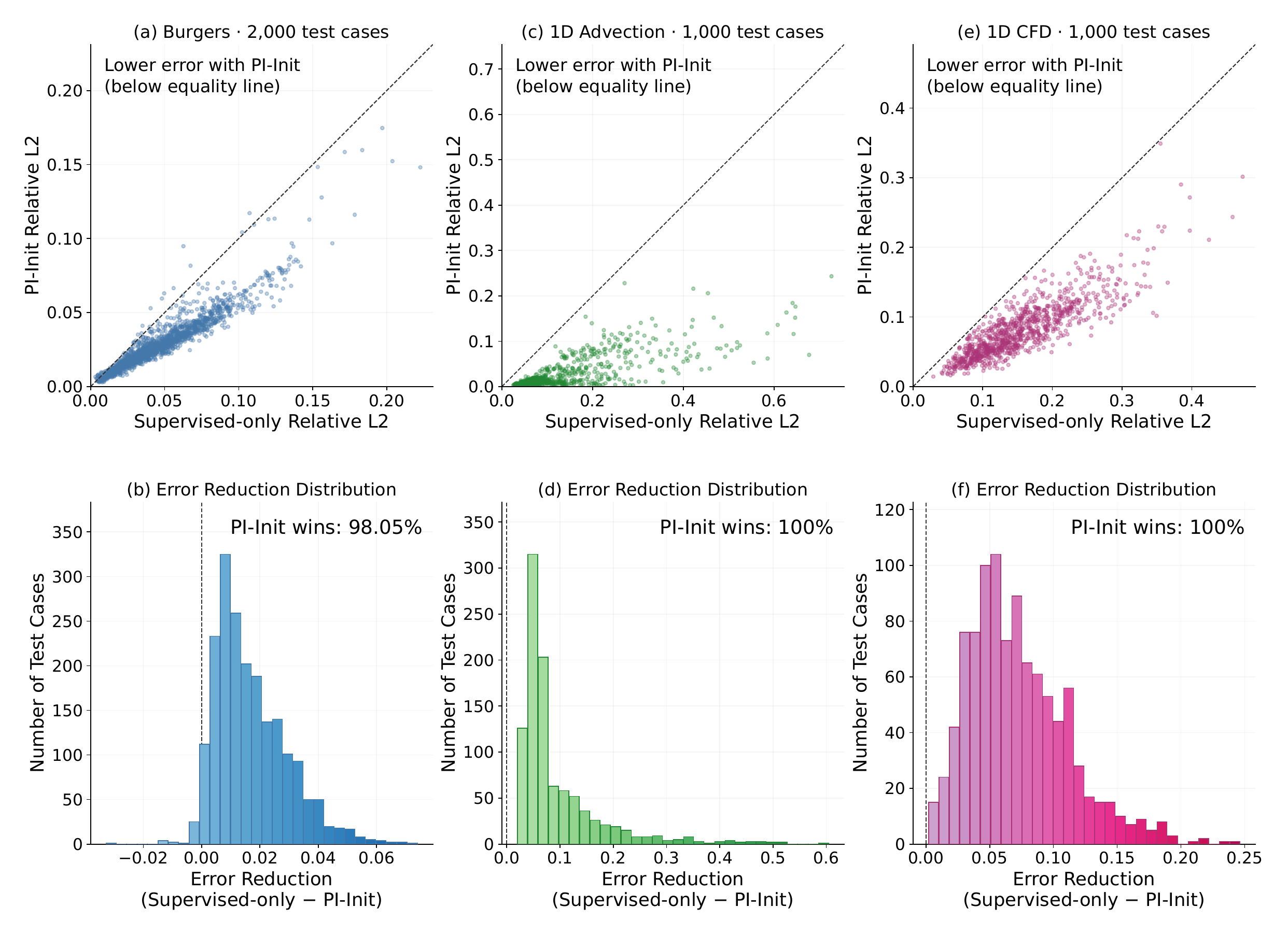}
\caption{Per-trajectory relative $L_2$ errors on Burgers, 1D Advection, and 1D CFD. Points below the identity line and positive histogram values favor \method{}. }
\label{fig:test-trajectory-distributions}
\end{figure}

\paragraph{Overall performance.}
We next evaluate whether retaining physics-pretrained parameters improves predictive accuracy compared with supervised-only training under matched labeling budgets and supervised training configurations. Table~\ref{tab:cross_architecture_comparison} demonstrates that \method{} consistently improves predictive accuracy across diverse PDE benchmarks and neural operator architectures. For FNO, prediction errors are reduced by 53.3\% on Burgers, 37.3\% on 1D CFD, 62.7\% on 1D Advection, and 10.2\% on 2D Navier--Stokes. Beyond accuracy, \method{} also reduces variability across random seeds, with the standard deviation decreasing from $0.0081$ to $0.0031$ on Burgers, from $0.0490$ to $0.0008$ on 1D Advection, and from $0.0088$ to $0.0009$ on 1D CFD. These observations are consistent with the initialization-variance contraction described in Corollary~\ref{cor:exact}(iii). Since both methods use identical labeled subsets and supervised training configurations, the results highlight the benefits of physics-informed initialization. Furthermore, Fig.~\ref{fig:test-trajectory-distributions} shows that most test trajectories achieve lower prediction errors with \method{}, with error-reduction distributions concentrated at positive values. This indicates that the improvements extend across the test sets rather than being driven by a few favorable examples.

\begin{figure}[pos=htbp]
\centering
\includegraphics[width=0.95\textwidth]{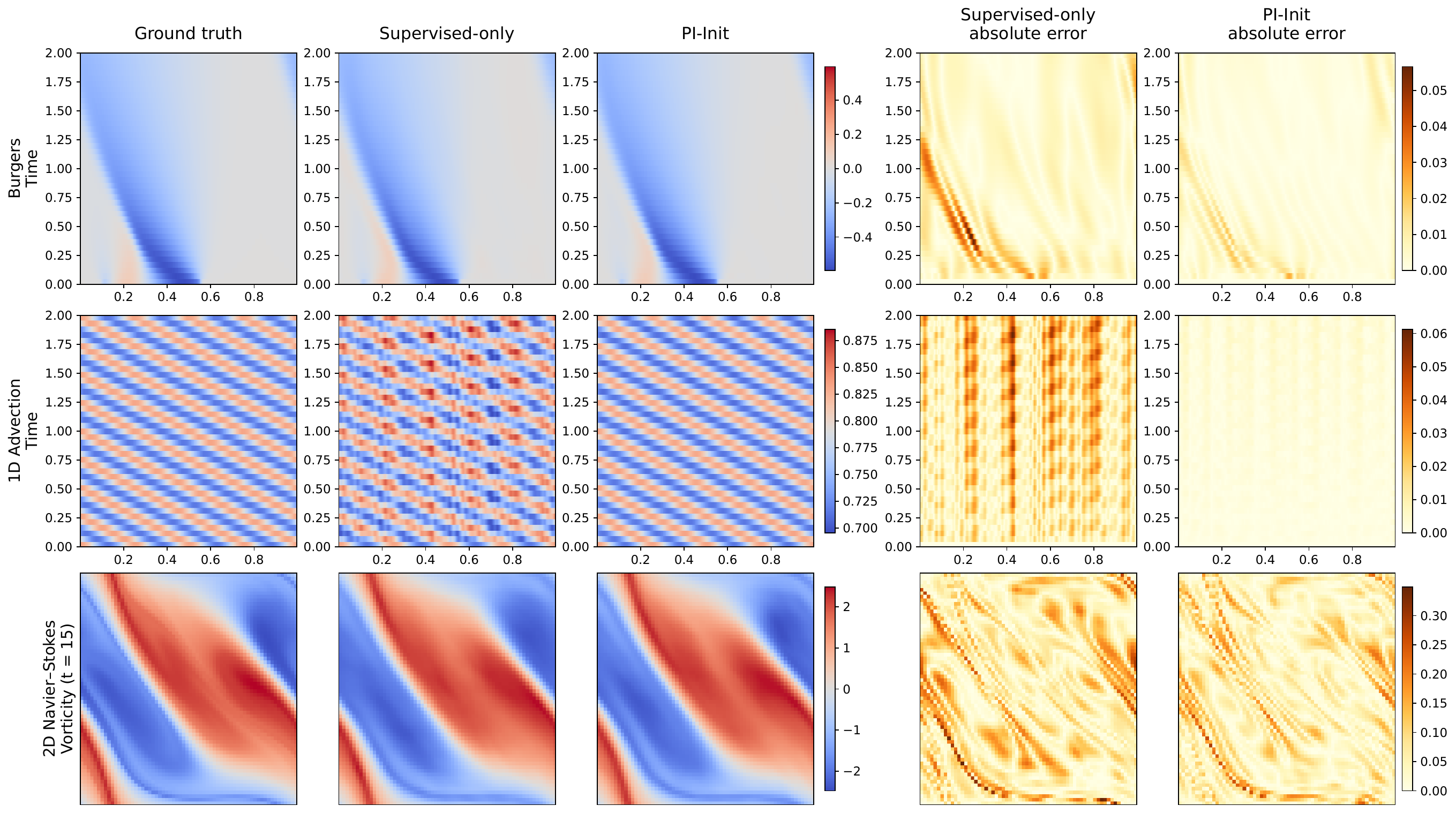}
\caption{FNO predictions on Burgers, 1D Advection, and 2D Navier–Stokes. Columns show the ground truth, supervised-only and \method{} predictions, and their absolute errors. Field and error maps use consistent color scales within each row.}
\label{fig:fno-field-comparison}
\end{figure}

\paragraph{Cross-architecture generalization.}
The improvements extend beyond FNO to WNO and LNO. \method{} reduces prediction errors by 40.0\% for WNO on 1D Reaction--Diffusion and 25.0\% for LNO on 1D Advection. On AirfRANS, all three architectures benefit, with test error reductions of 9.7\%, 6.1\%, and 6.6\% for FNO, WNO, and LNO, respectively. On 1D CFD, FNO achieves a 37.3\% reduction, whereas WNO and LNO exhibit smaller improvements of 0.9\% and 1.5\%. These results suggest that physics-informed initialization is effective across different operator architectures, including tasks where pretraining relies on approximate physical constraints. For 2D problems, PI10 and PI50 reduce FNO errors on Navier--Stokes by 7.4\% and 10.2\%, respectively, while additional experiments on 2D Reaction--Diffusion show improvements across all tested label budgets (Appendix~\ref{app:2d-results}). Qualitatively, Fig.~\ref{fig:fno-field-comparison} shows that \method{} produces predictions that more closely match the reference solutions, with smaller absolute errors across representative test trajectories. Additional spatial profiles are provided in Appendix~\ref{app:prediction-visualizations}.

\subsection{Label and Computational Efficiency}
\label{sec:label-efficiency}

\begin{figure}[pos=htbp]
\centering
\includegraphics[width=0.95\textwidth]{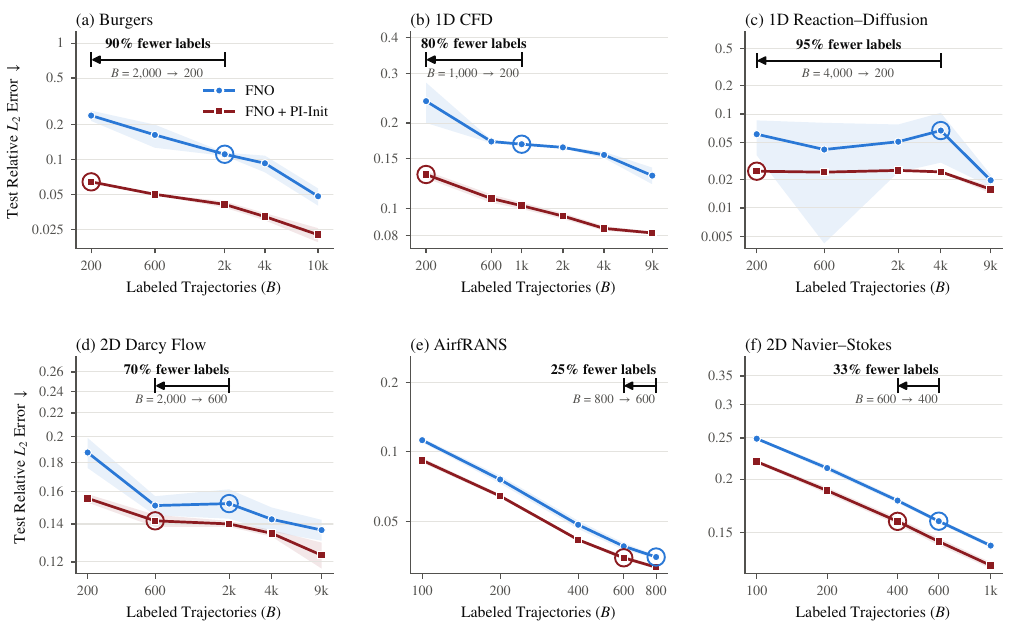}
\caption{Label efficiency across six PDE benchmarks (mean $\pm$ SD over three seeds). All datasets report test relative $L_2$ error except AirfRANS, which reports test error. Markers indicate label savings at comparable performance over the evaluated budgets, without interpolation.}
\label{fig:label-efficiency}
\end{figure}

\paragraph{Label efficiency.}
We next examine whether physics-informed initialization
can reduce the amount of labeled data required for
accurate prediction.
Theorem~\ref{thm:decomp} indicates that initialization has a greater influence in directions weakly constrained by labeled data, suggesting that physics-informed pretraining can be particularly beneficial under limited supervision. Fig.~\ref{fig:label-efficiency} compares \method{} with supervised-only training across six benchmarks and varying label budgets, while keeping the cheap pretraining inputs fixed. \method{} consistently achieves lower prediction errors across the evaluated budgets. Notably, \method{} achieves lower error on Burgers with only 200 labeled trajectories than supervised-only training with 2,000, corresponding to 90\% fewer labels. Similar savings are observed on 1D CFD (80\%), 2D Darcy Flow (70\%), AirfRANS (25\%), and 2D Navier--Stokes (approximately 33\%). On 1D Reaction--Diffusion, supervised-only training requires substantially more labels to reach comparable performance. Although the performance gap generally narrows as the label budget increases, these results demonstrate that physics-informed initialization can substantially reduce the need for expensive solution labels.

\paragraph{Cost.}
Beyond label efficiency, we also examine the computational overhead associated with physics-informed pretraining. Table~\ref{tab:compute-cost} reports the training time on Burgers and 1D CFD under the full-label setting. Physics-informed pretraining time increases approximately linearly with both the number of cheap inputs and the number of pretraining epochs. With $N=10{,}000$ unlabeled PDE inputs, the additional pretraining cost accounts for approximately 1.4--8.1\% of supervised training time for 500 epochs, increasing to 6.6--39.8\% with $N=50{,}000$, depending on the dataset and pretraining duration. These overheads are estimated relative to the measured supervised-only training times. Since \method{} retains the original architecture and inference operations, it introduces no additional inference complexity.

\begin{table}[pos=htbp]
\centering
\small
\setlength{\tabcolsep}{7pt}
\caption{Physics-informed pretraining and supervised-only training times (s) on 1D CFD and Burgers.}
\label{tab:compute-cost}
\begin{adjustbox}{max width=\linewidth}
\begin{tabular*}{\linewidth}{@{\extracolsep{\fill}}llccc@{}}
\toprule
Dataset & Cheap inputs $N$
& Pretraining (10 epochs)
& Pretraining (50 epochs)
& Supervised (500 epochs) \\
\midrule
\multirow{2}{*}{1D CFD ($B=9000$)}
& $10{,}000$ & $16.01\pm0.62$ & $77.82\pm3.55$
& \multirow{2}{*}{$1128.7\pm170.4$} \\
& $50{,}000$ & $73.97\pm1.78$ & $361.24\pm0.94$ & \\
\midrule
\multirow{2}{*}{Burgers ($B=10{,}000$)}
& $10{,}000$ & $31.17\pm0.37$ & $153.18\pm2.58$
& \multirow{2}{*}{$1902.8\pm10.6$} \\
& $50{,}000$ & $151.39\pm7.42$ & $757.15\pm29.57$ & \\
\bottomrule
\end{tabular*}
\end{adjustbox}
\end{table}

\subsection{Generalization Across OOD Scenarios}
\label{sec:res-ood}

To assess out-of-distribution (OOD) generalization, we evaluate
\method{} and supervised-only FNO models without further adaptation.
For Burgers, we vary the initial-condition distribution through
amplitude scaling, higher-frequency components, their combination,
and Gaussian random fields (GRFs), while keeping the governing
equation, viscosity ($\nu=0.01/\pi$), boundary conditions, and
numerical solver fixed. All models are evaluated using relative
$L_2$ errors over $t>0$. Table~\ref{tab:burgers-ood-config}
summarizes the initial-condition distributions.

\begin{table}[pos=htbp]
\centering
\small
\caption{Initial-condition distributions for Burgers ID and OOD evaluation.}
\label{tab:burgers-ood-config}
\begin{adjustbox}{max width=\linewidth}
\begin{tabular}{ll}
\toprule
Distribution & Initial-condition construction \\
\midrule
In-distribution (ID)
& $u_0^{\mathrm{test}}=u_0$ \\

Amplitude shift
& $u_0^{\mathrm{test}}=3u_0$ \\

Frequency shift
& $\displaystyle
u_0^{\mathrm{test}}=u_0^{\mathrm{HF}}
=\sum_{j=1}^{2}a_j\sin(k_jx+\phi_j),
\quad k_j\in[9,12]$ \\

Combined shift
& $u_0^{\mathrm{test}}=2u_0^{\mathrm{HF}}$ \\

GRF shift
& $u_0^{\mathrm{test}}\sim\mathcal{GP}(0,K_\ell),
\quad \ell=0.05,\ \sigma=0.55$ \\
\bottomrule
\end{tabular}
\end{adjustbox}
\end{table}

\begin{figure}[pos=htbp]
\centering
\includegraphics[width=0.99\linewidth]{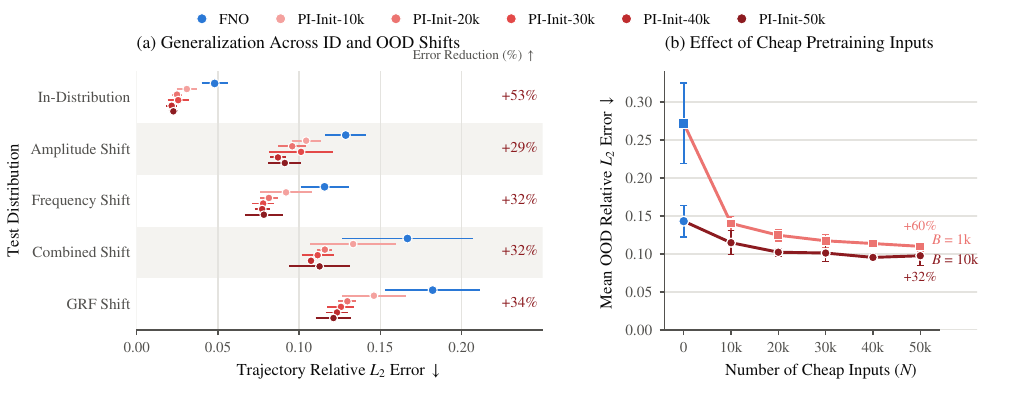}
\caption{OOD generalization on Burgers. (a) Test relative $L_2$
errors across ID and OOD scenarios with $B=10{,}000$. (b) Mean error
across four OOD scenarios versus the number of cheap pretraining
inputs $N$, for two labeling budgets.}
\label{fig:burgers-ood-cheap-inputs}
\end{figure}

As shown in Fig.~\ref{fig:burgers-ood-cheap-inputs}(a),
\method{} achieves lower mean relative $L_2$ errors across all
evaluated OOD shifts and the in-distribution (ID) test set, even
with the full labeled budget ($B=10{,}000$).
Fig.~\ref{fig:burgers-ood-cheap-inputs}(b) further shows that
increasing the number of cheap pretraining inputs $N$ generally
improves OOD performance, particularly under limited supervision.
Since larger $N$ also increases the number of optimization updates
at fixed pretraining epochs, the observed gains reflect the combined
effects of additional inputs and training.

Similar improvements are observed on 2D Navier--Stokes, with error
reductions of approximately 12--14\% across amplitude, frequency,
and GRF shifts, compared with 10\% on ID data
(Appendix~\ref{app:2d-results}). Overall, these results show improved
OOD generalization without additional solution labels or changes to
the operator architecture.

\subsection{Ablation Studies}
\label{sec:res-mechanism}


These ablation studies investigate which physical constraints provide useful initializations for supervised learning. We compare pretraining using only the PDE residual loss (PDE-only), only the boundary-condition residual loss (BC-only), and the full physics-informed objective in Eq.~\eqref{eq:pi_loss} (Full PI). We also compare physics-only predictions with predictions after supervised fine-tuning to assess the additional benefit of learning from solution labels. Experiments are conducted on Burgers and 2D Darcy Flow using the full labeling budgets and two pretraining input budgets ($N=10{,}000$ and $50{,}000$). The results are reported in Table~\ref{tab:ablation}.

\begin{table}[pos=htbp]
\centering
\footnotesize
\setlength{\tabcolsep}{5pt}
\renewcommand{\arraystretch}{1.08}
\caption{Ablation results on Burgers ($B=10{,}000$) and Darcy Flow ($B=9{,}000$). Test relative $L_2$ errors are reported as mean $\pm$ SD over three seeds. PI-only models are evaluated after 50 epochs of physics-informed training, while \method{} models undergo an additional 500 epochs of supervised fine-tuning. The supervised-only baseline uses 500 epochs of supervised training. $N$ denotes the number of unlabeled inputs used for pretraining. Lower is better.}
\label{tab:ablation}
\begin{adjustbox}{max width=\linewidth}
\begin{tabular}{lcccc}
\toprule
& \multicolumn{2}{c}{Burgers} & \multicolumn{2}{c}{Darcy Flow} \\
\cmidrule(lr){2-3}\cmidrule(lr){4-5}
Method & $N=10k$ & $N=50k$ & $N=10k$ & $N=50k$ \\
\midrule
Supervised-only & \multicolumn{2}{c}{$0.0480 \pm 0.0081$} & \multicolumn{2}{c}{$0.1366 \pm 0.0058$} \\
\midrule
\multicolumn{5}{l}{\textit{PI-only}} \\
\quad Full PI & $0.1544 \pm 0.0600$ & $0.1182 \pm 0.0207$ & $0.5689 \pm 0.0237$ & $0.4525 \pm 0.0227$ \\
\quad PDE-only & $0.2148 \pm 0.0825$ & $0.1277 \pm 0.0408$ & $0.8829 \pm 0.3726$ & $0.8835 \pm 0.3328$ \\
\midrule
\multicolumn{5}{l}{\textit{\method{} (with fine-tuning)}} \\
\quad Full PI & $0.0307 \pm 0.0060$ & $0.0224 \pm 0.0031$ & $\mathbf{0.1299 \pm 0.0013}$ & $\mathbf{0.1233 \pm 0.0065}$ \\
\quad PDE-only & $\mathbf{0.0298 \pm 0.0046}$ & $\mathbf{0.0193 \pm 0.0017}$ & $0.1306 \pm 0.0020$ & $0.1236 \pm 0.0107$ \\
\quad BC-only & $0.9999 \pm 0.0000$ & $0.9999 \pm 0.0000$ & $0.6448 \pm 0.0115$ & $0.6482 \pm 0.0063$ \\
\bottomrule
\end{tabular}
\end{adjustbox}
\end{table}

\paragraph{Pretraining and fine-tuning.}
We first examine how supervised fine-tuning affects the performance of physics-pretrained models. Table~\ref{tab:ablation} demonstrates that retaining physics-pretrained parameters for supervised fine-tuning is substantially more effective than using the physics-trained model directly. On Burgers, full \method{} reduces the supervised-only error from $0.0480$ to $0.0307$ and $0.0224$ with $N=10{,}000$ and $50{,}000$, corresponding to reductions of 36.0\% and 53.3\%, respectively. On Darcy Flow, the corresponding reductions are 4.9\% and 9.7\%. In contrast, PI-only predictions remain considerably less accurate than supervised-only training. These results indicate that the primary benefit of physics-informed pretraining lies in providing a transferable initialization.

\paragraph{Physics loss components.}
The PDE residual plays an important role in obtaining useful initializations for supervised learning. On Burgers, PDE-only pretraining yields a lower mean error after fine-tuning than Full PI, reaching $0.0193$ compared with $0.0224$ at $N=50{,}000$. On Darcy Flow, Full PI and PDE-only pretraining achieve similar fine-tuned errors ($0.1233$ and $0.1236$, respectively). Including boundary-condition constraints improves the pretrained Darcy Flow model's direct predictions, reducing the PI-only error from $0.8835$ to $0.4525$ at $N=50{,}000$. In contrast, BC-only pretraining results in higher errors after fine-tuning on both benchmarks. These results indicate that PDE-only pretraining followed by supervised fine-tuning already improves on supervised-only training. For Darcy Flow, adding the boundary-condition loss improves predictions before fine-tuning but provides little additional benefit after fine-tuning.

\paragraph{Joint versus staged training.}

\begin{figure}[pos=htbp]
\centering
\includegraphics[width=\linewidth]{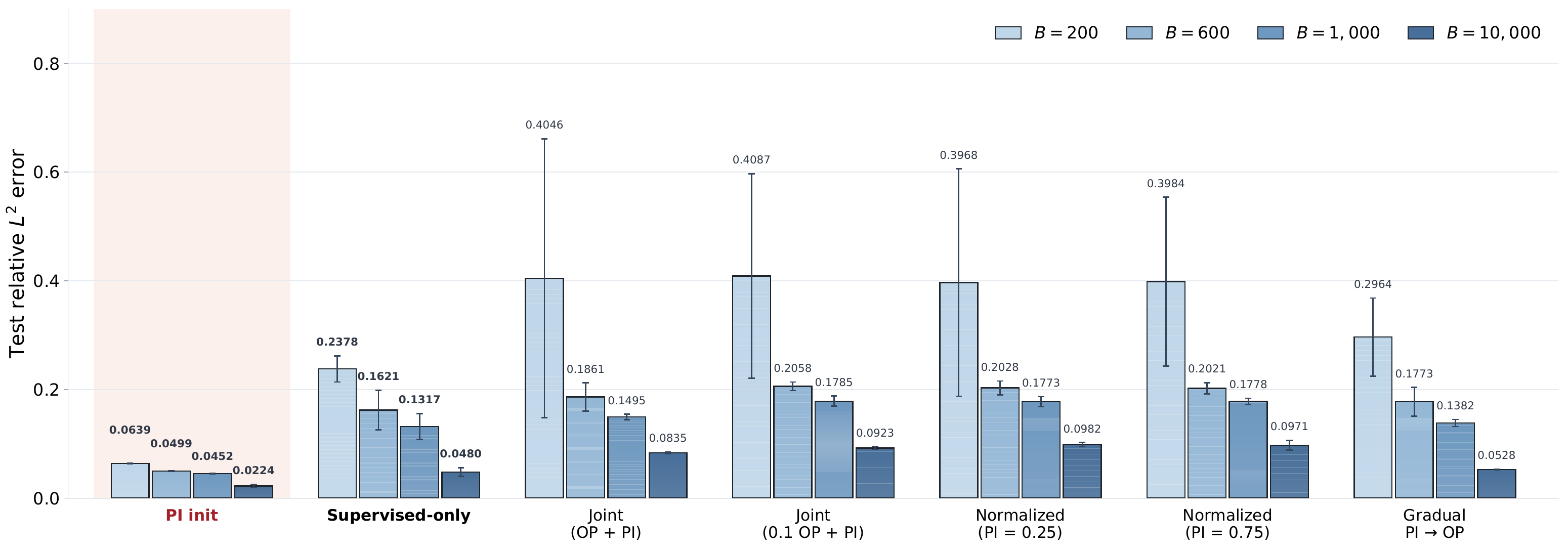}
\caption{Comparison of training strategies on Burgers across four label budgets ($B=200$, $600$, $1000$, and $10{,}000$). OP and PI denote supervised and physics-informed losses. Joint (OP + PI) and Joint (0.1 OP + PI) use the indicated loss weights; normalized variants apply initial loss-scale normalization with PI weights of 0.25 and 0.75; gradual PI$\rightarrow$OP progressively shifts from physics-informed to supervised optimization. Bars show mean test relative $L_2$ errors with standard deviations.}
\label{fig:report-joint}
\end{figure}

To assess the benefit of separating physics-informed pretraining from supervised fine-tuning, we compare \method{} with supervised-only training and five alternative physics--supervised training strategies inspired by PINO~\citep{pino}. The alternatives include unweighted and reweighted joint losses, normalized objectives with two different physics weights, and a gradual transition from physics-informed to supervised optimization. Fig.~\ref{fig:report-joint} shows that \method{} achieves the lowest mean test error across four labeling budgets on Burgers. At $B=1000$, \method{} achieves $0.0452$, compared with $0.1317$ for supervised-only training and $0.1382$ for the gradual schedule. At $B=10{,}000$, the corresponding errors are $0.0224$, $0.0480$, and $0.0528$. Under the tested configurations, none of the five alternatives achieves a lower mean error than supervised-only training. These results favor separate pretraining and fine-tuning stages over the evaluated joint and gradual training strategies on Burgers.

\section{Conclusion}
\label{sec:conclusion}

This study revisited the use of cheap physics for data-efficient neural operator learning, contrasting data selection with model initialization. The results show that physics-guided acquisition does not necessarily outperform random sampling, motivating a shift from selecting data to shaping the learner. Accordingly, Physics-Informed Initialization (\method{}) was proposed to transfer physics-pretrained parameters to downstream supervised learning. Theoretical analysis explains how such initialization can reduce prediction risk in directions weakly constrained by labeled data. Experiments across eight PDE benchmarks and three neural operator architectures demonstrate improvements in predictive accuracy, OOD generalization, and label efficiency. These findings establish physics-informed initialization as a practical approach to leveraging cheap physical information without additional solution labels. Future work will explore its extension to more complex PDE systems.

\section*{CRediT authorship contribution statement}
\textbf{Shuo Zhang:} Conceptualization, Methodology, Formal analysis, Software, Investigation, Visualization, Writing -- original draft, Writing -- review \& editing.
\textbf{Senwei Liang:} Conceptualization, Methodology, Supervision, Writing -- review \& editing.

\section*{Declaration of competing interest}
The authors declare that they have no known competing financial interests or personal relationships that could have appeared to influence the work reported in this paper.

\section*{Data availability}

The datasets used in this study are publicly available.
PDEBench is available at
\url{https://doi.org/10.18419/darus-2986}.
AirfRANS is available at
\url{https://github.com/Extrality/airfrans_lib}.
The 2D Navier--Stokes benchmark is available at
\url{https://github.com/meliao/fourier_neural_operator}.
The code to reproduce the results in this paper is available at \url{https://github.com/ShuoZHANG-BOT/CheapPhysics-NO}.

\section*{Declaration of generative AI and AI-assisted technologies in the writing process}
During the preparation of this work, the authors used generative AI tools to improve the language and readability of the manuscript. After using these tools, the authors reviewed and edited the content as needed and take full responsibility for the content of the publication.

\section*{Acknowledgments}
S. L. acknowledges partial support from start-up funds provided by Texas Tech University.

\bibliographystyle{cas-model2-names}
\bibliography{references}

\appendix
\renewcommand{\theproposition}{A.\arabic{proposition}}
\renewcommand{\thecorollary}{A.\arabic{corollary}}
\section{Supporting Theory and Proofs}
\label{app:theory-proofs}

This appendix provides the proofs and extensions of Section~\ref{sec:theory}. Appendix~\ref{app:notation} supplies the auxiliary facts. Appendices~\ref{app:exact-consequences} and~\ref{app:fidelity-analysis} develop the transfer consequences, and Appendix~\ref{app:cheap-input-analysis} examines the number of physics inputs.

\subsection{Notation and Auxiliary Facts}
\label{app:notation}

For a symmetric positive semidefinite (PSD) matrix $M$, $\|x\|_M^2=x^\top Mx$, and $\lambda_{\min}(M)$ and $\lambda_{\max}(M)$ denote its extreme eigenvalues. The notation is $\Phi_i=\Phi(a_i)$, $\xi_i$ for the label noise of sample $i$, and $\theta^\star$ for the target parameter of Assumption~\ref{ass:lm}. Noise variables of different samples are independent conditional on the inputs, so $\E[\xi_i\xi_j^\top\mid a_1,\ldots,a_B]=\delta_{ij}\,\omega^2(a_i)I_m$. For a fixed labeled set $S$ with $|S|=B$, define
\begin{equation}
H_S=\sum_{i\in S}\Phi_i^\top\Phi_i=B\Ah_S,
\qquad
b=\frac1B\sum_{i\in S}\Phi_i^\top\xi_i .
\label{eq:app-Hb}
\end{equation}

\begin{lemma}[Trace form of a quadratic risk]
\label{lem:trace}
If $z$ is a random vector with $\E[zz^\top]=C$, then $\E\|z\|_\Sigma^2=\tr(\Sigma C)$.
\end{lemma}
\begin{proof}
$\E\|z\|_\Sigma^2=\E\,\tr(z^\top\Sigma z)=\E\,\tr(\Sigma zz^\top)=\tr(\Sigma\,\E[zz^\top])$, by linearity of trace and expectation.
\end{proof}

\begin{lemma}[Arithmetic--harmonic mean inequality for traces]
\label{lem:amhm}
For any symmetric positive definite $M\in\mathbb R^{d\times d}$, $\tr(M)\,\tr(M^{-1})\ge d^2$.
\end{lemma}
\begin{proof}
Let $\lambda_1,\ldots,\lambda_d>0$ be the eigenvalues of $M$. By the Cauchy--Schwarz inequality,
$d^2=\bigl(\sum_j\sqrt{\lambda_j}\cdot\lambda_j^{-1/2}\bigr)^2\le\bigl(\sum_j\lambda_j\bigr)\bigl(\sum_j\lambda_j^{-1}\bigr)=\tr(M)\tr(M^{-1})$.
\end{proof}

\begin{lemma}[Matrix Chernoff inequality; \citealp{tropp}, Thm.~1.1]
\label{lem:chernoff}
Let $X_1,\ldots,X_B$ be independent random PSD $d\times d$ matrices with $\lambda_{\max}(X_i)\le R$ almost surely, and let $\mu_{\min}=\lambda_{\min}(\sum_i\E X_i)$ and $\mu_{\max}=\lambda_{\max}(\sum_i\E X_i)$. Then, for $\varepsilon\in(0,1)$,
\begin{align}
\Pr\Bigl[\lambda_{\min}\Bigl(\sum_iX_i\Bigr)\le(1-\varepsilon)\mu_{\min}\Bigr]&\le d\Bigl[\frac{e^{-\varepsilon}}{(1-\varepsilon)^{1-\varepsilon}}\Bigr]^{\mu_{\min}/R}\le d\,e^{-\varepsilon^2\mu_{\min}/(2R)},\\
\Pr\Bigl[\lambda_{\max}\Bigl(\sum_iX_i\Bigr)\ge(1+\varepsilon)\mu_{\max}\Bigr]&\le d\Bigl[\frac{e^{\varepsilon}}{(1+\varepsilon)^{1+\varepsilon}}\Bigr]^{\mu_{\max}/R}\le d\,e^{-\varepsilon^2\mu_{\max}/(3R)}.
\end{align}
\end{lemma}

\begin{lemma}[Gradient descent on a quadratic]
\label{lem:gd}
Let $f(\theta)=\frac12(\theta-\bar\theta)^\top H(\theta-\bar\theta)-g^\top(\theta-\bar\theta)$ with $H$ symmetric PSD, and run $\theta_{t+1}=\theta_t-\eta\nabla f(\theta_t)$ from $\theta_0$. Then
\begin{equation}
\theta_t-\bar\theta=(I-\eta H)^t(\theta_0-\bar\theta)+\eta\sum_{j=0}^{t-1}(I-\eta H)^j g .
\end{equation}
If $0<\eta\le1/\lambda_{\max}(H)$, then $0\preceq I-\eta H\preceq I$.
\end{lemma}
\begin{proof}
Since $\nabla f(\theta)=H(\theta-\bar\theta)-g$, the error $\epsilon_t=\theta_t-\bar\theta$ satisfies $\epsilon_{t+1}=(I-\eta H)\epsilon_t+\eta g$. Unrolling this recursion $t$ times gives the formula. Every eigenvalue of $I-\eta H$ has the form $1-\eta\lambda$ with $\lambda\in[0,\lambda_{\max}(H)]$, and it lies in $[0,1]$ when $\eta\lambda_{\max}(H)\le1$.
\end{proof}

\begin{lemma}[Two scalar facts]
\label{lem:scalar}
(a) For $x\ge0$ and $\gamma>0$, $x\,e^{-\gamma x}\le1/(e\gamma)$, with equality at $x=1/\gamma$.
(b) If $X\sim\mathcal N(\mu,\tau^2)$, then $\operatorname{Var}(X^2)=2\tau^4+4\mu^2\tau^2$.
\end{lemma}
\begin{proof}
(a) The derivative of $x e^{-\gamma x}$ is $(1-\gamma x)e^{-\gamma x}$, which vanishes only at $x=1/\gamma$, where the function attains its maximum $1/(e\gamma)$.
(b) Write $X=\mu+\tau Z$ with $Z\sim\mathcal N(0,1)$. Then $X^2=\mu^2+2\mu\tau Z+\tau^2Z^2$. Using $\E Z=\E Z^3=0$, $\E Z^2=1$, and $\E Z^4=3$, the moments are $\E X^2=\mu^2+\tau^2$ and $\E X^4=\mu^4+6\mu^2\tau^2+3\tau^4$, so $\operatorname{Var}(X^2)=\E X^4-(\E X^2)^2=2\tau^4+4\mu^2\tau^2$.
\end{proof}

\subsection{Selection Risk under Bounded Leverage}
\label{app:proof-random}

The leverage $\ell(a)=\tr(\Phi(a)\Sigma^{-1}\Phi(a)^\top)$ measures the information carried by an input relative to the test distribution, with $\E_\mu\ell(a)=d$. Bounded leverage means
\begin{equation}
\ell(a)\le\kappa d\quad\mu\text{-almost surely},\qquad \kappa\ge1.
\label{eq:bounded-leverage}
\end{equation}
Pool inputs lie in the support of $\mu$, and Random samples from $\mu$.

\begin{proposition}[Selection risk under bounded leverage]
\label{prop:selection-risk}
Under Assumption~\ref{ass:lm} and Eq.~\eqref{eq:bounded-leverage}, let $\omega^2(a)\equiv\sigma^2$ and let $\hat\theta_S$ be the least-squares fit on $S$. (i) For every subset of $B$ pool inputs that is chosen from the inputs alone (independently of the label noise) and has a full-rank design, $\Ah_S\succ0$,
\begin{equation}
\E_\xi\,\mathcal E(\hat\theta_S)\;\ge\;\frac{\sigma^2 d}{\kappa B}.
\end{equation}
(ii) If $S_{\rm rand}$ consists of $B$ i.i.d.\ draws from $\mu$ and $B\ge 3\kappa d\log(2d/\delta)/\varepsilon^2$ for $\varepsilon\in(0,1)$, then with probability at least $1-\delta$ the design is full rank and
\begin{equation}
\frac{\sigma^2 d}{(1+\varepsilon)B}\;\le\;\E_\xi\,\mathcal E(\hat\theta_{S_{\rm rand}})\;\le\;\frac{\sigma^2 d}{(1-\varepsilon)B}.
\end{equation}
Consequently,
$\E_\xi\mathcal E(\hat\theta_{S_{\rm rand}})\le\frac{\kappa}{1-\varepsilon}\min_{|S|=B,\,\Ah_S\succ0}\E_\xi\mathcal E(\hat\theta_S)$,
so no full-rank selector, however it uses physics, can reduce Random's least-squares risk by more than a fraction $1-(1-\varepsilon)/\kappa$.
\end{proposition}

\begin{proof}
The least-squares risk follows from its noise covariance. Part (i) concerns full-rank designs, so $H_S=B\Ah_S$ is invertible and the least-squares estimator is unique. Then $\hat\theta_S=H_S^{-1}\sum_{i\in S}\Phi_i^\top u_i$. Substituting $u_i=\Phi_i\theta^\star+\xi_i$ gives
\begin{equation}
\hat\theta_S-\theta^\star=H_S^{-1}\sum_{i\in S}\Phi_i^\top\xi_i .
\end{equation}
Because $S$ is chosen from the inputs alone, conditioning on $S$ leaves the noise distribution unchanged; by the independence of the noise across samples,
\begin{equation}
\E_\xi\bigl[(\hat\theta_S-\theta^\star)(\hat\theta_S-\theta^\star)^\top\bigr]
=H_S^{-1}\Bigl(\sum_{i\in S}\Phi_i^\top\sigma^2I_m\Phi_i\Bigr)H_S^{-1}=\sigma^2H_S^{-1}.
\end{equation}
Lemma~\ref{lem:trace} then gives $\E_\xi\mathcal E(\hat\theta_S)=\sigma^2\tr(\Sigma H_S^{-1})$.

To relate this risk to the leverage, let $M=\Sigma^{-1/2}H_S\Sigma^{-1/2}$. By cyclicity of the trace,
\begin{equation}
\tr(\Sigma H_S^{-1})=\tr(\Sigma^{1/2}H_S^{-1}\Sigma^{1/2})=\tr(M^{-1}),
\end{equation}
and
\begin{equation}
\tr(M)=\sum_{i\in S}\tr\bigl(\Sigma^{-1/2}\Phi_i^\top\Phi_i\Sigma^{-1/2}\bigr)=\sum_{i\in S}\tr\bigl(\Phi_i\Sigma^{-1}\Phi_i^\top\bigr)=\sum_{i\in S}\ell(a_i)\le B\kappa d .
\end{equation}

For the lower bound in (i), Lemma~\ref{lem:amhm} gives $\tr(M^{-1})\ge d^2/\tr(M)\ge d^2/(B\kappa d)=d/(\kappa B)$. Multiplying by $\sigma^2$ proves (i).

For the bounds in (ii), let $S_{\rm rand}=\{a_1,\ldots,a_B\}$ be i.i.d.\ draws from $\mu$ and $X_i=\Sigma^{-1/2}\Phi_i^\top\Phi_i\Sigma^{-1/2}$, so that $M=\sum_iX_i$. The $X_i$ are independent and PSD, $\E X_i=\Sigma^{-1/2}\Sigma\Sigma^{-1/2}=I$, and hence $\mu_{\min}=\mu_{\max}=B$. Moreover, by Eq.~\eqref{eq:bounded-leverage}, $\lambda_{\max}(X_i)\le\tr(X_i)=\ell(a_i)\le\kappa d=:R$ almost surely, as Lemma~\ref{lem:chernoff} requires. Lemma~\ref{lem:chernoff} gives
\begin{equation}
\Pr[\lambda_{\min}(M)\le(1-\varepsilon)B]\le d\,e^{-\varepsilon^2B/(2\kappa d)},
\qquad
\Pr[\lambda_{\max}(M)\ge(1+\varepsilon)B]\le d\,e^{-\varepsilon^2B/(3\kappa d)}.
\end{equation}
If $B\ge3\kappa d\log(2d/\delta)/\varepsilon^2$, each right-hand side is at most $\delta/2$. A union bound shows that, with probability at least $1-\delta$, every eigenvalue of $M$ lies in $[(1-\varepsilon)B,(1+\varepsilon)B]$. In particular, $M$ is invertible, and
\begin{equation}
\frac{d}{(1+\varepsilon)B}\le\tr(M^{-1})\le\frac{d}{(1-\varepsilon)B}.
\end{equation}
Multiplying by $\sigma^2$ proves (ii).

Finally, dividing the upper bound in (ii) by the lower bound in (i) gives
\begin{equation}
\E_\xi\mathcal E(\hat\theta_{S_{\rm rand}})\le\frac{\kappa}{1-\varepsilon}\min_{|S|=B,\,\Ah_S\succ0}\E_\xi\mathcal E(\hat\theta_S).
\end{equation}
Equivalently, the best possible relative reduction is $1-\min_S\E_\xi\mathcal E(\hat\theta_S)/\E_\xi\mathcal E(\hat\theta_{S_{\rm rand}})\le1-(1-\varepsilon)/\kappa$.
\end{proof}

\subsection{Prediction Difficulty and Acquisition Utility}
\label{app:proof-difficulty}

Candidate difficulty and acquisition utility describe different quantities. In a scalar-output linear-Gaussian model, let $\theta\mid\mathcal D\sim\mathcal N(\bar\theta,P)$ and $u=\phi^\top\theta+\xi_a$, with independent noise $\xi_a\sim\mathcal N(0,\omega^2(a))$. Define
\begin{equation}
D(a)=\E[(u-\phi^\top\bar\theta)^2]=\phi^\top P\phi+\omega^2(a)>0,
\qquad U(a)=\tr(\Sigma P)-\tr(\Sigma P_{+a}),
\end{equation}
where $P_{+a}$ is the posterior covariance after observing the candidate label.

\begin{proposition}[Difficulty is not utility]
\label{prop:difficulty}
Let $\lambda_{\min},\lambda_{\max}$ be the extreme eigenvalues of $P^{1/2}\Sigma P^{1/2}$. Then
\begin{equation}
U(a)=\frac{\phi^\top P\Sigma P\phi}{D(a)},
\qquad
\lambda_{\min}\Big(1-\frac{\omega^2(a)}{D(a)}\Big)\le U(a)\le\lambda_{\max}\Big(1-\frac{\omega^2(a)}{D(a)}\Big).
\end{equation}
In particular: (i) for fixed $\phi$, increasing the irreducible error $\omega^2(a)$ \emph{increases} difficulty and \emph{decreases} utility; (ii) for deterministic labels ($\omega\equiv0$), $U$ is invariant to rescaling $\phi$, so the residual magnitude carries no information about utility beyond the direction of $P^{1/2}\phi$.
\end{proposition}

\begin{proof}
The residual of the current posterior mean is $u-\phi^\top\bar\theta=\phi^\top(\theta-\bar\theta)+\xi$. The two terms are independent with zero mean, so $D=\E[(u-\phi^\top\bar\theta)^2]=\phi^\top P\phi+\omega^2$.

To compute the utility, note that the pair $(\theta,u)$ is jointly Gaussian with $\operatorname{Cov}(\theta,u)=P\phi$ and $\operatorname{Var}(u)=D$. Gaussian conditioning (the Schur complement) gives
\begin{equation}
P_{+a}=\operatorname{Cov}(\theta\mid u)=P-\frac{P\phi\phi^\top P}{D}.
\end{equation}
This expression does not depend on the realized value of $u$, so the utility is a deterministic function of $a$. The division is well defined because $D>0$ by assumption, which also covers $\omega=0$ whenever $\phi^\top P\phi>0$.

The Bayes risk of the posterior mean under the test metric is $\E[\|\theta-\bar\theta\|_\Sigma^2]=\tr(\Sigma P)$ by Lemma~\ref{lem:trace}. Therefore,
\begin{equation}
U=\tr(\Sigma P)-\tr(\Sigma P_{+a})=\frac{\tr(\Sigma P\phi\phi^\top P)}{D}=\frac{\phi^\top P\Sigma P\phi}{D}.
\end{equation}

For the two-sided bounds, let $z=P^{1/2}\phi$ and $K=P^{1/2}\Sigma P^{1/2}$. Then $\phi^\top P\Sigma P\phi=z^\top Kz$, and the Rayleigh quotient gives $\lambda_{\min}\|z\|^2\le z^\top Kz\le\lambda_{\max}\|z\|^2$. Since $\|z\|^2=\phi^\top P\phi=D-\omega^2$, dividing by $D$ yields
$\lambda_{\min}(1-\omega^2/D)\le U\le\lambda_{\max}(1-\omega^2/D)$.

The two consequences now follow directly. (i) For fixed $\phi$, $D=\phi^\top P\phi+\omega^2$ increases with $\omega^2$, whereas $U=\phi^\top P\Sigma P\phi/(\phi^\top P\phi+\omega^2)$ has a fixed numerator and therefore decreases.
(ii) With $\omega=0$ and any scalar $c\ne0$, $U(c\phi)=c^2\phi^\top P\Sigma P\phi/(c^2\phi^\top P\phi)=U(\phi)$. Hence $U$ depends only on the direction of $P^{1/2}\phi$, while $D(c\phi)=c^2D(\phi)$ can be made arbitrarily large.
\end{proof}

\subsection{Proof of Theorem~\ref{thm:decomp}}
\label{app:proof-decomp}

\emph{Claim.} The expected fine-tuned risk splits into $\|P_T(\theta_{\rm init}-\theta^\star)\|_\Sigma^2$ plus a term independent of $\theta_{\rm init}$, and the component of the error in $\ker\Ah_S$ never changes.

\begin{proof}
The fine-tuning objective has a quadratic form. Let $\epsilon=\theta-\theta^\star$. Since $u_i=\Phi_i\theta^\star+\xi_i$,
\begin{equation}
F_B(\theta)=\frac1{2B}\sum_{i\in S}\|\Phi_i\epsilon-\xi_i\|^2
=\frac12\epsilon^\top\Ah_S\epsilon-b^\top\epsilon+\frac1{2B}\sum_{i\in S}\|\xi_i\|^2,
\end{equation}
with $b$ as in Eq.~\eqref{eq:app-Hb}. This is the quadratic of Lemma~\ref{lem:gd} with $H=\Ah_S$, $\bar\theta=\theta^\star$, and $g=b$.

Consequently, by Lemma~\ref{lem:gd}, $\epsilon_T=\theta_T-\theta^\star$ satisfies
\begin{equation}
\epsilon_T=P_T\epsilon_0+Q_Tb,
\qquad
Q_T=\eta\sum_{j=0}^{T-1}(I-\eta\Ah_S)^j,
\qquad
\epsilon_0=\theta_{\rm init}-\theta^\star .
\label{eq:app-closed}
\end{equation}

The information boundary removes the cross term between noise and initialization. Expanding the risk gives
\begin{equation}
\mathcal E(\theta_T)=\|P_T\epsilon_0\|_\Sigma^2+2\langle P_T\epsilon_0,\Sigma Q_Tb\rangle+\|Q_Tb\|_\Sigma^2 .
\end{equation}
Condition on the inputs. By Assumption~\ref{ass:boundary}, $\epsilon_0$ (through $\theta_{\rm init}$) and $S$ are independent of the noise $\{\xi_i\}$. Because $\E[\xi_i\mid a_i]=0$, it follows that $\E[b\mid\text{inputs},\theta_{\rm init}]=0$. Hence
$\E\bigl[\langle P_T\epsilon_0,\Sigma Q_Tb\rangle\mid\text{inputs},\theta_{\rm init}\bigr]=\langle P_T\epsilon_0,\Sigma Q_T\,\E[b\mid\text{inputs},\theta_{\rm init}]\rangle=0$, and the tower property removes the cross term.

The first term does not involve the noise and equals $\mathcal B_T(\theta_{\rm init})$. The second, $V_T(S)=\E_\xi\|Q_Tb\|_\Sigma^2$, depends only on $S$, $\eta$, $T$, and the noise law, not on $\theta_{\rm init}$. With $\omega\equiv\sigma$, $\operatorname{Cov}(b)=\sigma^2\Ah_S/B$, and Lemma~\ref{lem:trace} gives $V_T(S)=(\sigma^2/B)\tr(\Sigma Q_T\Ah_SQ_T)$.

It remains to prove the null-space statement. Let $\Pi_0$ be the orthogonal projector onto $\ker\Ah_S$. Because $\Ah_S$ is symmetric, $\Ah_S\Pi_0=0$ and $\Pi_0\Ah_S=(\Ah_S\Pi_0)^\top=0$. Hence $\Pi_0(I-\eta\Ah_S)=\Pi_0$, which gives $\Pi_0P_T=\Pi_0$ and $\Pi_0Q_T=\eta T\Pi_0$. Moreover, $b\in\operatorname{span}\bigcup_i\operatorname{range}(\Phi_i^\top)=\operatorname{range}(\Ah_S)=(\ker\Ah_S)^\perp$, so $\Pi_0b=0$. Applying $\Pi_0$ to Eq.~\eqref{eq:app-closed} gives $\Pi_0\epsilon_T=\Pi_0\epsilon_0$ for every $T$. Finally, $\dim\ker\Ah_S\ge d-\operatorname{rank}(H_S)\ge d-mB$, since $H_S$ is a sum of $B$ matrices of rank at most $m$.
\end{proof}

\subsection{Proof of Theorem~\ref{thm:modewise}}
\label{app:proof-modewise}

\emph{Claim.} Under spectral compatibility, $\Delta_{\rm init}=\sum_k\alpha_k[e_k^2-(c_ke_k+(1-c_k)\beta_k)^2]$; this gain is nonnegative for matched physics, and its limit as $T'\to\infty$ yields the fidelity boundary~\eqref{eq:fidelity-boundary}.

\begin{proof}
The PI loss $\widehat\Lpi(\theta)=\frac12(\theta-\tilde\theta)^\top H_{\rm PI}(\theta-\tilde\theta)+\text{const}$ is the quadratic of Lemma~\ref{lem:gd} with $g=0$. Running $T'$ steps from $\theta_0$ therefore gives $\theta_{\rm PI}-\tilde\theta=C(\theta_0-\tilde\theta)$ with $C=(I-\eta'H_{\rm PI})^{T'}$. Adding and subtracting $\theta^\star$,
\begin{equation}
\theta_{\rm PI}-\theta^\star=C(\theta_0-\theta^\star)+(I-C)(\tilde\theta-\theta^\star).
\end{equation}

Under Assumption~\ref{ass:sc}, $Cv_k=c_kv_k$ with $c_k=(1-\eta'\rho_k)^{T'}$. Taking the inner product with $v_k$ gives
\begin{equation}
\langle\theta_{\rm PI}-\theta^\star,v_k\rangle=c_ke_k+(1-c_k)\beta_k .
\end{equation}
The minimizer $\tilde\theta$ is unique only up to $\ker H_{\rm PI}$. This ambiguity affects $\beta_k$ only on modes with $\rho_k=0$, where $c_k=1$ and the coefficient $1-c_k$ vanishes, so the expression is well defined.

For the bias, since $P_Tv_k=(1-\eta s_k)^Tv_k$ and $\Sigma v_k=w_kv_k$, for any $x=\sum_kx_kv_k$,
\begin{equation}
\|P_Tx\|_\Sigma^2=\sum_kw_k(1-\eta s_k)^{2T}x_k^2=\sum_k\alpha_kx_k^2 .
\end{equation}

The parameters $\theta_{\rm PI}$ use only inputs and the equation, so they satisfy Assumption~\ref{ass:boundary}, and Theorem~\ref{thm:decomp} applies to both arms with the same $S$, $\eta$, and $T$. The variance terms $V_T(S)$ are therefore identical and cancel. Using the eigenbasis expression above with $x_k=e_k$ for Ordinary and $x_k=c_ke_k+(1-c_k)\beta_k$ for \method{} gives Eq.~\eqref{eq:modewise}.

For matched physics, $\beta_k=0$ reduces the $k$th summand to $\alpha_k(1-c_k^2)e_k^2$, which is nonnegative because $\alpha_k\ge0$ and $c_k\in[0,1]$.

For the limit, Assumption~\ref{ass:sc} gives $0<\eta'\rho_k\le1$ whenever $\rho_k>0$, so $c_k\to0$ as $T'\to\infty$ and the $k$th summand tends to $\alpha_k(e_k^2-\beta_k^2)$. Modes with $\rho_k=0$ have $c_k=1$ for every $T'$ and contribute zero. Summing over $k$ gives $\Delta_{\rm init}^{\infty}=\sum_{k:\rho_k>0}\alpha_k(e_k^2-\beta_k^2)$, which is positive exactly when inequality~\eqref{eq:fidelity-boundary} holds.
\end{proof}

\subsection{Consequences of Matched Physics}
\label{app:exact-consequences}

Under Assumptions~\ref{ass:lm}--\ref{ass:sc}, the mode-wise formula yields duration, variance, and distribution-shift consequences for matched physics.

\begin{corollary}[Matched physics]
\label{cor:exact}
If $\beta_k=0$ for all $k$, then $\Delta_{\rm init}=\sum_k\alpha_k(1-c_k^2)e_k^2$, and:
\begin{enumerate}
\item[(i)] \textbf{Monotone duration.} $\Delta_{\rm init}$ is nondecreasing in $T'$ and, for a fixed step size $\eta'$, in every $\rho_k$.
\item[(ii)] \textbf{Vanishing gain with supervised optimization.} If $w_k\le C_w s_k$ whenever $\alpha_k>0$, then $\Delta_{\rm init}\le C_w\|\theta_0-\theta^\star\|^2/(2e\,\eta T)$, i.e., the gain decays at least as $O(1/T)$ in the number $T$ of full-batch gradient steps. Conversely, if $\ker\Ah_S$ carries test mass, $\Delta_{\rm init}\ge\sum_{k:s_k=0}w_k(1-c_k^2)e_k^2$, which does not vanish as $T$ grows.
\item[(iii)] \textbf{Initialization-induced variance contraction.} If $\theta_0\sim\mathcal N(0,\tau^2I)$, then $\operatorname{Var}_{\theta_0}[\mathcal B_T]=\sum_k\alpha_k^2\gamma_k^4(2\tau^4+4\tau^2\langle\theta^\star,v_k\rangle^2)$ with $\gamma_k=1$ for ordinary training and $\gamma_k=c_k\le1$ for \method{}. This variance is taken over random initializations with the data and optimization dynamics fixed.
\item[(iv)] \textbf{Spectrally compatible distribution shift.} Suppose the shifted test law leaves $\theta^\star$ unchanged and its feature covariance $\Sigma^{\rm ood}$ shares the eigenbasis $\{v_k\}$, with eigenvalues $w_k^{\rm ood}$. Then $\Delta_{\rm init}^{\rm ood}=\sum_kw_k^{\rm ood}(1-\eta s_k)^{2T}(1-c_k^2)e_k^2\ge0$, and shifts that move mass toward weakly constrained modes that the PI inputs excite ($\rho_k>0$) receive the largest absolute gains.
\end{enumerate}
\end{corollary}

\begin{proof}
Set $\beta_k=0$. By Theorem~\ref{thm:modewise}, the $k$th summand of $\Delta_{\rm init}$ is $\alpha_k(e_k^2-c_k^2e_k^2)=\alpha_k(1-c_k^2)e_k^2$.

\emph{(i) Monotone duration.}
For fixed $\eta'$, $c_k=(1-\eta'\rho_k)^{T'}\in[0,1]$ is nonincreasing in $T'$ and in $\rho_k$, so every summand $\alpha_k(1-c_k^2)e_k^2$ is nondecreasing in both.

\emph{(ii) Decay with the number of fine-tuning steps.}
Because $0\le\eta s_k\le1$, it follows that $0\le1-\eta s_k\le e^{-\eta s_k}$, so $(1-\eta s_k)^{2T}\le e^{-2\eta Ts_k}$. If $\alpha_k>0$, the hypothesis $w_k\le C_ws_k$ and Lemma~\ref{lem:scalar}(a) with $\gamma=2\eta T$ give
\begin{equation}
\alpha_k\le C_w\,s_ke^{-2\eta Ts_k}\le\frac{C_w}{2e\eta T}.
\end{equation}
Hence $\Delta_{\rm init}\le\sum_k\alpha_ke_k^2\le\frac{C_w}{2e\eta T}\sum_ke_k^2=\frac{C_w\|\theta_0-\theta^\star\|^2}{2e\eta T}$. For the lower bound, modes with $s_k=0$ have $\alpha_k=w_k$, and all other summands are nonnegative.

\emph{(iii) Initialization-induced variance.}
If $\theta_0\sim\mathcal N(0,\tau^2I)$, the coordinates $\langle\theta_0,v_k\rangle$ are i.i.d.\ $\mathcal N(0,\tau^2)$, because an orthonormal change of basis preserves an isotropic Gaussian. Hence the $e_k=\langle\theta_0,v_k\rangle-\langle\theta^\star,v_k\rangle\sim\mathcal N(-\langle\theta^\star,v_k\rangle,\tau^2)$ are independent. The contractions $c_k$ are deterministic, so $\mathcal B_T=\sum_k\alpha_k\gamma_k^2e_k^2$ with $\gamma_k=1$ (Ordinary) or $\gamma_k=c_k$ (\method{}). By independence and Lemma~\ref{lem:scalar}(b),
\begin{equation}
\operatorname{Var}_{\theta_0}[\mathcal B_T]=\sum_k\alpha_k^2\gamma_k^4\operatorname{Var}(e_k^2)=\sum_k\alpha_k^2\gamma_k^4\bigl(2\tau^4+4\tau^2\langle\theta^\star,v_k\rangle^2\bigr),
\end{equation}
and the \method{} value is at most the Ordinary value term by term, since $c_k\le1$.

\emph{(iv) Shifted test law.}
The training quantities $s_k$, $c_k$, and $e_k$ are fixed when the test law changes. Under each test law, both initializations share the same label-variance term, which cancels in their risk difference. If the shift leaves $\theta^\star$ unchanged and the shifted covariance $\Sigma^{\rm ood}$ shares the eigenbasis with eigenvalues $w_k^{\rm ood}\ge0$, the last two parts of the proof of Theorem~\ref{thm:modewise} hold verbatim with $w_k$ replaced by $w_k^{\rm ood}$. Each summand remains nonnegative, and the gain on mode $k$ is proportional to $w_k^{\rm ood}(1-\eta s_k)^{2T}(1-c_k^2)$. It is therefore largest where the shifted law places mass on modes with small $s_k$ and large $\rho_k$.
\end{proof}

\paragraph{Spectral compatibility under distribution shift.}
The shared eigenbasis in Corollary~\ref{cor:exact}(iv) fixes the sign of transfer mode by mode. A covariance that couples modes can change the comparison. For post-training errors $(1,1)$ under ordinary training and $(0,1)$ under \method{}, $\Sigma=I$ gives risks $2$ and $1$, respectively. Under $\Sigma^{\rm ood}=\bigl(\begin{smallmatrix}1&-0.9\\-0.9&1\end{smallmatrix}\bigr)$, the corresponding risks are $0.2$ and $1$.

\subsection{Physics Mismatch and Pretraining Duration}
\label{app:fidelity-analysis}

Under mismatch, the post-pretraining error of a mode determines the preferred pretraining duration. Under Assumptions~\ref{ass:lm}--\ref{ass:sc}, consider a mode with $\rho_k>0$ and $\beta_k\ne0$, and let $q=1-\eta'\rho_k\in(0,1)$, so that $c_k=q^{T'}$ decreases from $1$ (no pretraining) to $0$ (convergence). The post-pretraining error $\epsilon_k(c)=\beta_k+c\,(e_k-\beta_k)$ satisfies:
\begin{enumerate}
\item[(i)] if $\beta_ke_k<0$, some finite pretraining duration achieves strictly smaller error than pretraining to convergence, and longer pretraining eventually increases the error;
\item[(ii)] if $\beta_ke_k>0$ and $|\beta_k|<|e_k|$, pretraining to convergence is optimal and reduces the error by the factor $|\beta_k/e_k|$;
\item[(iii)] if $\beta_ke_k\ge0$ and $|\beta_k|\ge|e_k|$, no pretraining duration improves the mode, and every duration with $c_k<1$ is strictly worse when $|\beta_k|>|e_k|$.
\end{enumerate}

\begin{proof}
Drop the index $k$. The error $\epsilon(c)$ is affine in $c$, with $\epsilon(1)=e$ and $\epsilon(0)=\beta$.
(i) If $\beta e<0$, then $\epsilon$ vanishes at $c^\star=\beta/(\beta-e)\in(0,1)$ and $|\epsilon(c)|=|e-\beta|\,|c-c^\star|$. Choose the finite $T'$ with $q^{T'}\in(qc^\star,c^\star]$; its error is at most $|e-\beta|(1-q)c^\star=(1-q)|\beta|<|\beta|$, the error at convergence. For $q^{T'}<c^\star$, $|\epsilon|=|e-\beta|(c^\star-q^{T'})$ increases with $T'$.
(ii)--(iii) If $\beta e\ge0$, then $|\epsilon(c)|=(1-c)|\beta|+c|e|$ is a convex combination of $|\beta|$ and $|e|$. It is minimized at $c=0$ when $|\beta|<|e|$, and it is at least $|e|$ for every $c$ when $|\beta|\ge|e|$, with strict inequality for $c<1$ when $|\beta|>|e|$.
\end{proof}

\subsection{Staged Retention and Joint Training}
\label{app:proof-joint}

\begin{proposition}[Staged retention versus joint training]
\label{prop:joint}
Under Assumptions~\ref{ass:lm}--\ref{ass:sc} with $\omega\equiv\sigma$, $\lambda>0$, and $0<\eta s_k\le1$, consider a mode with $s_k>0$ and $\rho_k>0$, and let $b=B^{-1}\sum_{i\in S}\Phi(a_i)^\top\xi_i$ with $b_k=\langle b,v_k\rangle$. As $T\to\infty$, staged \method{} converges to the least-squares error $b_k/s_k$, with $\E b_k^2=\sigma^2s_k/B$, independently of the physics. Minimizing the joint objective $F_B+\lambda\widehat\Lpi$ gives the error $(b_k+\lambda\rho_k\beta_k)/(s_k+\lambda\rho_k)$. Joint training has lower expected error on mode $k$ if and only if
\begin{equation}
\beta_k^2<\frac{\sigma^2(2s_k+\lambda\rho_k)}{B\,s_k\,\lambda\rho_k}.
\end{equation}
In particular, with deterministic solver labels ($\sigma\to0$), joint training is asymptotically biased on every label-identified mode with $\beta_k\ne0$ for every $\lambda>0$, whereas staged retention is not.
\end{proposition}

\begin{proof}
By the quadratic form of $F_B$ derived in the proof of Theorem~\ref{thm:decomp} and Assumption~\ref{ass:sc}, the fine-tuning objective restricted to mode $k$ is $\frac{s_k}2\epsilon_k^2-b_k\epsilon_k$ (up to a constant), where $b_k=\langle b,v_k\rangle$. With $\omega\equiv\sigma$, $\operatorname{Cov}(b)=\sigma^2\Ah_S/B$, so $\E b_k=0$ and $\E b_k^2=\sigma^2v_k^\top\Ah_Sv_k/B=\sigma^2s_k/B$. The PI loss restricted to mode $k$ is $\frac{\rho_k}2(\epsilon_k-\beta_k)^2$.

Consider first staged retention. Gradient descent on the fine-tuning objective gives $\epsilon_{t+1,k}=(1-\eta s_k)\epsilon_{t,k}+\eta b_k$. For $0<\eta s_k\le1$, this converges to the unique fixed point $b_k/s_k$ from any initialization, including $\theta_{\rm PI}$. The asymptotic expected squared error is therefore $\E b_k^2/s_k^2=\sigma^2/(Bs_k)$, independent of the physics.

For joint training, the objective on mode $k$ is $\frac{s_k}2\epsilon_k^2-b_k\epsilon_k+\frac{\lambda\rho_k}2(\epsilon_k-\beta_k)^2$. Setting the derivative $s_k\epsilon_k-b_k+\lambda\rho_k(\epsilon_k-\beta_k)$ to zero gives the minimizer $\epsilon_k^{\rm J}=(b_k+\lambda\rho_k\beta_k)/(s_k+\lambda\rho_k)$. Because $\beta_k$ depends only on inputs and the equation and $\E b_k=0$,
\begin{equation}
\E(\epsilon_k^{\rm J})^2=\frac{\sigma^2s_k/B+\lambda^2\rho_k^2\beta_k^2}{(s_k+\lambda\rho_k)^2}.
\end{equation}

Comparing the two, joint training is better on mode $k$ if and only if $\E(\epsilon_k^{\rm J})^2<\sigma^2/(Bs_k)$. Multiplying by $(s_k+\lambda\rho_k)^2>0$ gives
\begin{equation}
\lambda^2\rho_k^2\beta_k^2<\frac{\sigma^2}{Bs_k}\bigl[(s_k+\lambda\rho_k)^2-s_k^2\bigr]=\frac{\sigma^2}{Bs_k}\lambda\rho_k(2s_k+\lambda\rho_k).
\end{equation}
For $\rho_k>0$, dividing by $\lambda^2\rho_k^2$ yields the stated condition $\beta_k^2<\sigma^2(2s_k+\lambda\rho_k)/(Bs_k\lambda\rho_k)$. As $\sigma\to0$, the right-hand side vanishes, so joint training is strictly worse whenever $\beta_k\ne0$. Its error tends to $\lambda\rho_k\beta_k/(s_k+\lambda\rho_k)\ne0$, while the staged error tends to $0$.
\end{proof}

\subsection{Number of Physics Pretraining Inputs}
\label{app:cheap-input-analysis}

With matched, noise-free physics, the minimizers of the quadratic surrogate form $\theta^\star+\ker H_{\rm PI}^{(N)}$. For nested input sets, adding residual constraints shrinks this null space until the inputs span the population physics range. Once that range is covered, the eigenvalues $\rho_k^{(N)}$ still determine the contraction $c_k=(1-\eta'\rho_k^{(N)})^{T'}$. Thus the number of inputs affects both the directions constrained by physics and the conditioning of pretraining. At a fixed number of epochs and a fixed minibatch size, $T'$ also grows with the input count. The experiments varying $N$ consequently vary coverage and optimization effort together.

\subsection{Residual Stability for Burgers}
\label{app:burgers-energy}

This estimate shows that the residual-stability idea behind the linearized analysis holds for the viscous Burgers equation used in the experiments of Section~\ref{sec:experiments-main}.

\begin{proposition}[Residual--error estimate for viscous Burgers]
\label{prop:burgers-energy}
Let $u$ solve $u_t+uu_x=\nu u_{xx}$ on a periodic interval, let $\hat u$ be a sufficiently smooth periodic approximation with residual $r=\hat u_t+\hat u\hat u_x-\nu\hat u_{xx}$, and suppose $\|\hat u_x-\frac12u_x\|_{L^\infty}\le M$ on $[0,t]$. Then the error $e=\hat u-u$ satisfies
\begin{equation}
\|e(t)\|_{L^2}^2\le e^{(2M+1)t}\Bigl[\|e(0)\|_{L^2}^2+\int_0^t\|r(s)\|_{L^2}^2\,ds\Bigr].
\end{equation}
\end{proposition}

\begin{proof}
Subtracting the two equations and using the identity $\hat u\hat u_x-uu_x=ue_x+e\hat u_x$ gives
\begin{equation}
e_t+ue_x+e\hat u_x-\nu e_{xx}=r .
\end{equation}

Multiply by $e$ and integrate over the periodic domain; boundary terms vanish by periodicity. Integration by parts gives
\begin{equation}
\int ue_xe\,dx=\tfrac12\int u\,(e^2)_x\,dx=-\tfrac12\int u_xe^2\,dx,
\qquad
-\nu\int e_{xx}e\,dx=\nu\int e_x^2\,dx .
\end{equation}
Therefore,
\begin{equation}
\frac12\frac{d}{dt}\|e\|_{L^2}^2+\nu\|e_x\|_{L^2}^2+\int\Bigl(\hat u_x-\frac12u_x\Bigr)e^2\,dx=\int re\,dx .
\end{equation}

Next, drop the nonnegative viscous term. Bound $\int(\hat u_x-\frac12u_x)e^2\,dx\ge-M\|e\|_{L^2}^2$, and use Young's inequality $\int re\,dx\le\frac12\|r\|_{L^2}^2+\frac12\|e\|_{L^2}^2$. Then
\begin{equation}
\frac{d}{dt}\|e\|_{L^2}^2\le(2M+1)\|e\|_{L^2}^2+\|r\|_{L^2}^2 .
\end{equation}

Gr\"onwall's inequality completes the bound. With $\gamma=2M+1$, the previous inequality gives $\frac{d}{dt}\bigl(e^{-\gamma t}\|e\|_{L^2}^2\bigr)\le e^{-\gamma t}\|r\|_{L^2}^2$. Integrating from $0$ to $t$,
\begin{equation}
\|e(t)\|_{L^2}^2\le e^{\gamma t}\|e(0)\|_{L^2}^2+\int_0^te^{\gamma(t-s)}\|r(s)\|_{L^2}^2\,ds\le e^{\gamma t}\Bigl[\|e(0)\|_{L^2}^2+\int_0^t\|r(s)\|_{L^2}^2\,ds\Bigr].
\end{equation}
\end{proof}

When the initial condition is inserted exactly into the rollout, as in the implementation used here, $e(0)=0$, and the trajectory error is controlled entirely by the time-integrated residual. The implemented loss uses discrete derivatives, so a fully discrete version of this estimate additionally contains differentiation and quadrature errors. These errors are one source of the misspecification $\beta$ in Theorem~\ref{thm:modewise}.


\section{Additional Experimental Results}
\label{app:additional-results}

This appendix provides additional acquisition diagnostics, prediction visualizations, two-dimensional benchmark results, and checkpoint-transfer analysis.

\subsection{Diagnostics for Residual-Based Acquisition}
\label{app:selection-diagnostics}

\paragraph{Residual-based acquisition scores.}
Three acquisition scores are constructed from the PDE residual of a predicted solution $\hat{u}$. For the viscous Burgers equation, the residual is defined as
\begin{equation}
r(\hat{u})
=
\partial_t\hat{u}
+
\hat{u}\,\partial_x\hat{u}
-
\nu\,\partial_{xx}\hat{u}.
\end{equation}
Let $\langle\cdot\rangle$ denote averaging over the spatial grid and interior time points of a predicted trajectory. The absolute mean residual is defined as
\begin{equation}
s_{\mathrm{MAE,abs}}
=
\langle |r(\hat{u})|\rangle.
\end{equation}
To account for differences in the magnitudes of individual PDE terms, two normalized scores are also considered. Let $q_1=\partial_t\hat{u}$, $q_2=\hat{u}\,\partial_x\hat{u}$, and $q_3=\nu\,\partial_{xx}\hat{u}$. The normalized RMS and relative MAE scores are
\begin{align}
s_{\mathrm{RMS}}
&=
\frac{\sqrt{\langle r^2\rangle}}
{\max\left(\sum_{j=1}^{3}\sqrt{\langle q_j^2\rangle},\,10^{-8}\right)},\\
s_{\mathrm{MAE,rel}}
&=
\frac{\langle |r|\rangle}
{\max\left(\sum_{j=1}^{3}\langle |q_j|\rangle,\,10^{-8}\right)}.
\end{align}
All scores are computed from probe predictions without accessing candidate solution labels, using centered finite differences in time and FFT-based spectral differentiation in space. Normalized RMS PRE ($s_{\mathrm{RMS}}$) and Absolute MAE PRE ($s_{\mathrm{MAE,abs}}$) are used for acquisition in Table~\ref{tab:selection-mean}, while Relative MAE PRE ($s_{\mathrm{MAE,rel}}$) is included as an additional diagnostic.

The evaluated acquisition strategies follow methods proposed in previous studies~\citep{polanska2026,musekamp2025,holzmuller2023framework}. Normalized RMS PRE and Absolute MAE PRE combine 90 uniformly sampled candidates with 10 score-guided selections per round. PRE--SBAL and QbC--SBAL employ score-proportional sampling without replacement, based on mean absolute PDE residuals and predictive disagreement between two independently initialized models, respectively. LCMD performs greedy feature-space selection using representations extracted from the probe FNO.

\begin{figure}[pos=htbp]
\centering
\includegraphics[width=0.8\linewidth]{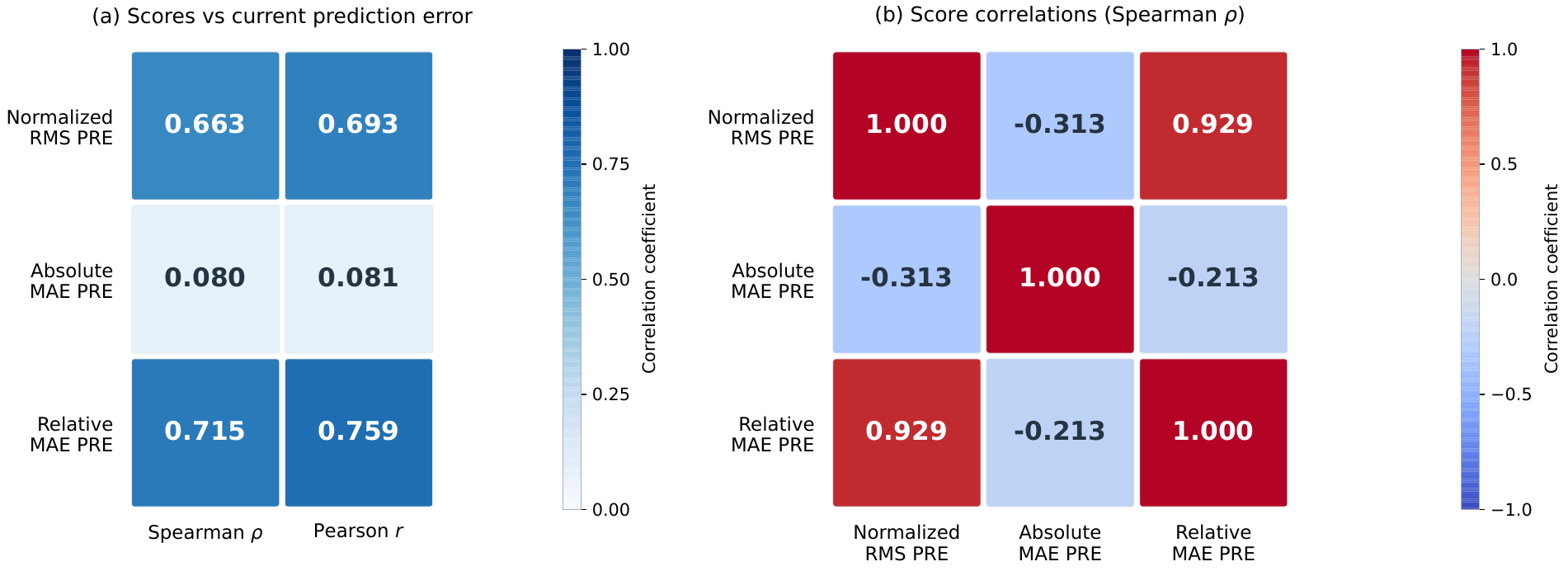}
\caption{Residual-score diagnostics on Burgers. (a) Correlations with probe prediction errors. (b) Spearman rank correlations between acquisition scores.}
\label{fig:selection-pre-correlation}
\end{figure}

\paragraph{Acquisition diagnostics.}
Fig.~\ref{fig:selection-pre-correlation} examines the relationships between residual-based scores, prediction errors, and candidate rankings. Across 960 pooled score--error pairs, Relative MAE PRE exhibits the strongest correlation with the current probe error (Spearman $\rho=0.715$, Pearson $r=0.759$), followed by Normalized RMS PRE ($\rho=0.663$, $r=0.693$). In contrast, Absolute MAE PRE shows little correlation ($\rho=0.080$, $r=0.081$). The two normalized scores also produce similar candidate rankings (Spearman $\rho=0.929$), whereas Absolute MAE PRE is negatively correlated with both normalized measures. Despite their ability to identify difficult predictions, these scores do not yield better acquired subsets than uniform random sampling under the evaluated protocols, reinforcing the distinction between prediction difficulty and labeling utility.

\subsection{Additional Prediction Visualizations}
\label{app:prediction-visualizations}

\begin{figure}[pos=htbp]
\centering
\includegraphics[width=0.9\linewidth]{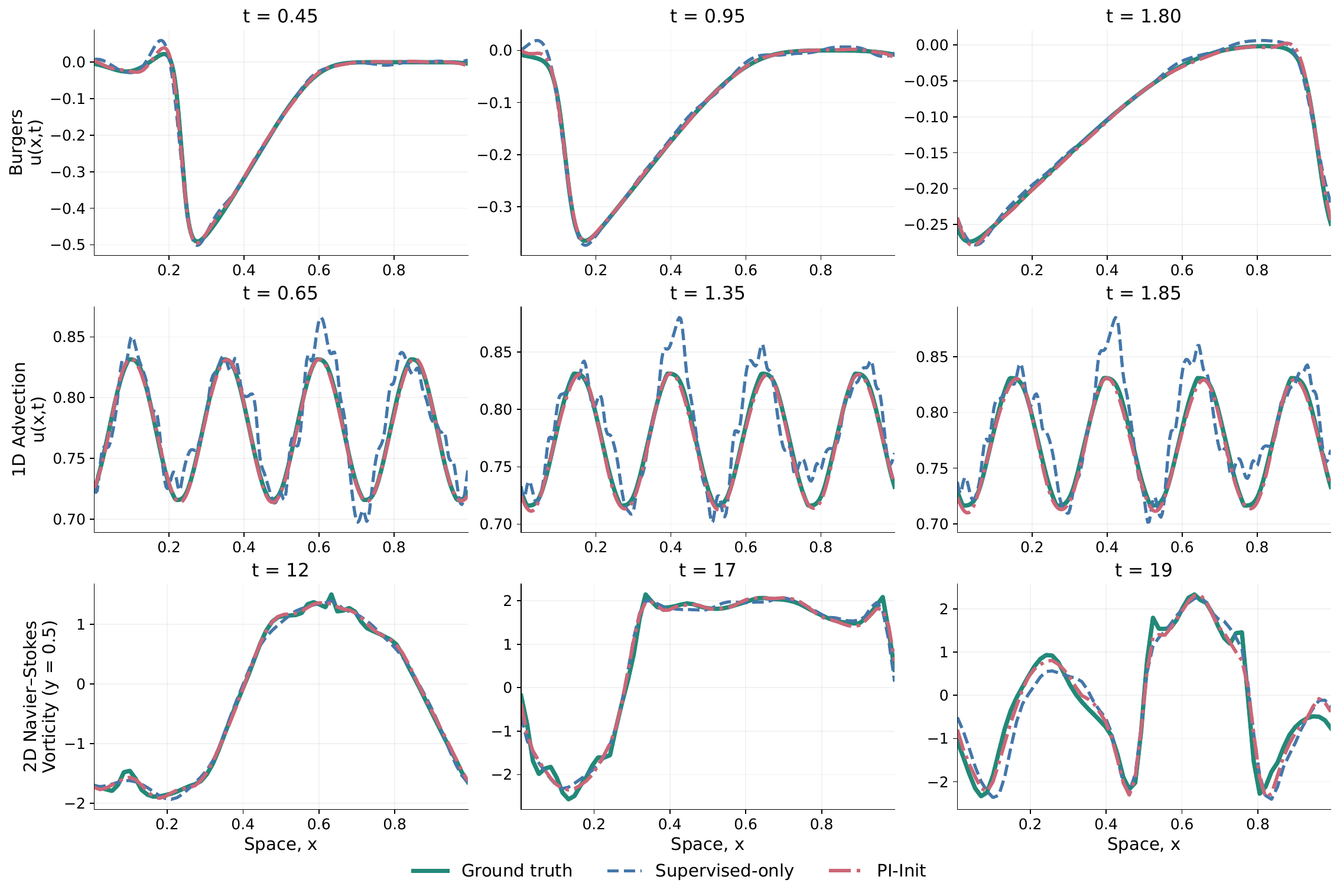}
\caption{Spatial profiles of representative test trajectories on Burgers, 1D Advection, and 2D Navier–Stokes at selected prediction times. Curves show the ground truth, supervised-only FNO, and \method{} predictions.}
\label{fig:fno-spatial-profiles}
\end{figure}

Fig.~\ref{fig:fno-spatial-profiles} provides additional spatial comparisons for the representative test trajectories shown in Fig.~\ref{fig:fno-field-comparison}. At selected prediction times, \method{} generally follows the reference solutions more closely than supervised-only training, further illustrating the improvements in prediction accuracy.

\subsection{Additional Two-Dimensional Results}
\label{app:2d-results}

\begin{figure}[pos=htbp]
\centering
\includegraphics[width=0.95\textwidth]{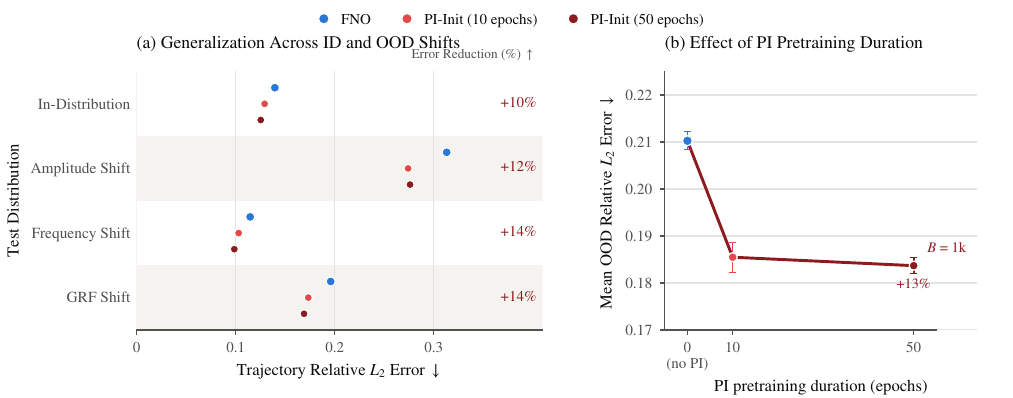}
\caption{OOD generalization on 2D Navier--Stokes. (a) Test relative $L_2$ errors across ID and OOD scenarios. (b) Mean OOD error versus physics-pretraining epochs. }
\label{fig:ns2d-ood}
\end{figure}

\paragraph{2D Navier--Stokes.}
Following the benchmark of \citet{fno}, experiments are conducted on a $64\times64$ periodic grid with viscosity $\nu=10^{-5}$, using 1,000 training and 200 test trajectories. The FNO receives the first 10 vorticity frames and predicts the following 10 frames. Physics-informed pretraining uses 50,000 unlabeled initial vorticity fields drawn from a Gaussian random field, followed by 500 epochs of supervised fine-tuning. To evaluate out-of-distribution (OOD) generalization, additional test sets of 200 trajectories are generated under amplitude scaling ($\times1.5$ and $\times2$), changes in Gaussian random field parameters ($\tau=3$ and $14$), and increased spectral roughness ($\alpha=2$). As shown in Fig.~\ref{fig:ns2d-ood}, \method{} achieves lower prediction errors across all evaluated shifts. The mean OOD error decreases from $0.210$ for supervised-only training to $0.185$ with PI10 and $0.184$ with PI50, demonstrating that the benefits of physics-informed initialization extend beyond the training distribution.

\paragraph{2D Reaction--Diffusion.}
The coupled 2D Reaction--Diffusion system is evaluated on $128\times128$ spatial grids with 101 time frames. The FNO receives 10 initial frames and predicts the remaining 91 frames. Four labeled budgets ($B\in\{200,400,600,900\}$) are evaluated using PI10 followed by 500 epochs of supervised fine-tuning. As shown in Table~\ref{tab:rd2d_pi10_ft500}, \method{} achieves lower mean prediction errors across all evaluated budgets, with the largest relative reduction at $B=600$. At the full training budget ($B=900$), the mean errors are similar, but \method{} exhibits substantially lower variability across random seeds.

\begin{table}[pos=htbp]
\centering
\small
\setlength{\tabcolsep}{6pt}
\caption{Test relative $L_2$ errors on 2D Reaction--Diffusion
(physics-informed pretraining for 10 epochs, followed by
supervised fine-tuning for 500 epochs). Values are reported
as mean $\pm$ standard deviation.}
\label{tab:rd2d_pi10_ft500}
\begin{adjustbox}{max width=\linewidth}
\begin{tabular}{rccc}
\toprule
Labels $B$ & Supervised-only & PI-Init + fine-tuning & Improvement (\%) \\
\midrule
200 & $0.1699 \pm 0.0136$ & $\mathbf{0.1619 \pm 0.0027}$ & 4.7 \\
400 & $0.1457 \pm 0.0028$ & $\mathbf{0.1359 \pm 0.0074}$ & 6.7 \\
600 & $0.1240 \pm 0.0070$ & $\mathbf{0.1121 \pm 0.0067}$ & 9.6 \\
900 & $0.0880 \pm 0.0165$ & $\mathbf{0.0872 \pm 0.0026}$ & 0.9 \\
\bottomrule
\end{tabular}
\end{adjustbox}
\end{table}

\subsection{PI Checkpoint Transfer}
\label{app:pi-checkpoint-transfer}

To investigate the relationship between pretraining duration and downstream performance, intermediate PI checkpoints are evaluated on Burgers with $N=50{,}000$ cheap inputs and $B=10{,}000$ labeled trajectories. Each checkpoint is fine-tuned for 500 supervised epochs under identical downstream settings. Table~\ref{tab:burgers-pi-trajectory} compares the standalone PI errors with the corresponding fine-tuned results.

\begin{table}[pos=htbp]
\centering
\small
\setlength{\tabcolsep}{6pt}
\caption{PI-Init checkpoint transfer on Burgers ($N=50{,}000$, $B=10{,}000$). Each PI-Init checkpoint is fine-tuned for 500 epochs. Values are test relative $L_2$ errors.}
\label{tab:burgers-pi-trajectory}
\begin{adjustbox}{max width=\linewidth}
\begin{tabular}{rcc}
\toprule
PI-Init epochs & Before fine-tuning & After 500-epoch fine-tuning \\
\midrule
0  & -- & $0.0480 \pm 0.0081$ \\
1  & $0.1995 \pm 0.0119$ & $0.0442 \pm 0.0006$ \\
5  & $0.1418 \pm 0.0144$ & $0.0308 \pm 0.0034$ \\
10 & $0.2110 \pm 0.1219$ & $0.0256 \pm 0.0006$ \\
20 & $0.1464 \pm 0.0302$ & $0.0375 \pm 0.0178$ \\
30 & $0.1100 \pm 0.0156$ & $\mathbf{0.0206 \pm 0.0023}$ \\
40 & $\mathbf{0.0990 \pm 0.0093}$ & $0.0224 \pm 0.0050$ \\
50 & $0.1182 \pm 0.0207$ & $0.0224 \pm 0.0031$ \\
\bottomrule
\end{tabular}
\end{adjustbox}
\end{table}

The lowest PI-only error occurs at epoch 40, whereas the best downstream performance is achieved with the epoch-30 checkpoint. This observation is consistent with Theorem~\ref{thm:modewise}, indicating that better standalone physics-informed predictions do not necessarily yield better initializations for supervised learning.

\section{Experimental Settings and Hyperparameters}
\label{app:experimental_details}

This section details the benchmark configurations, training protocols, and out-of-distribution (OOD) evaluation settings.

\subsection{Benchmarks and Physics Residuals}
\label{app:benchmarks}

Table~\ref{tab:benchmarks} summarizes the PDE inputs and physics-informed residuals used for pretraining. Residuals are classified as exact when they correspond to the data-generating equations up to numerical discretization, and approximate when simplified governing equations are employed.

\begin{table}[pos=htbp]
\centering
\small
\caption{Benchmark inputs and physics-informed residuals used for pretraining.}
\label{tab:benchmarks}
\begin{adjustbox}{max width=\linewidth}
\begin{tabular}{llll}
\toprule
Dataset & Input $a$ & PI residual & Fidelity \\
\midrule
Burgers & Initial condition & $u_t+uu_x-\nu u_{xx}$, periodic BC & Exact \\
1D Advection & Initial condition & $u_t+\beta u_x$, periodic BC & Exact \\
1D Reaction--Diffusion & Initial condition & $u_t-\nu u_{xx}-\rho u(1-u)$ & Exact \\
2D Reaction--Diffusion & Initial fields & FitzHugh--Nagumo system & Exact \\
Darcy Flow & Coefficient field & $-\nabla\cdot(a\nabla u)-f$, Dirichlet BC & Exact \\
Diffusion--Sorption & Initial state & Retardation--diffusion residual & Exact \\
2D Navier--Stokes & Initial vorticity & Vorticity-form residual, periodic BC & Exact \\
1D CFD & Conservative state & Near-inviscid Euler conservation & Approximate \\
AirfRANS & Geometry/conditions & Continuity + steady Euler momentum & Approximate \\
\bottomrule
\end{tabular}
\end{adjustbox}
\end{table}

\subsection{Training Configurations}
\label{app:training_protocols}

Table~\ref{tab:fno-training-config} summarizes the standard FNO training configurations. The supervised-only baseline is trained from random initialization, whereas \method{} initializes supervised fine-tuning with physics-pretrained parameters. Both methods use the same supervised training configuration. Table~\ref{tab:fno-data-counts} reports the corresponding pretraining input counts and labeled sample budgets. Task-specific deviations are described alongside the relevant experiments.

\begin{table}[pos=htbp]
\centering
\small
\setlength{\tabcolsep}{8pt}
\caption{Standard FNO training configurations.}
\label{tab:fno-training-config}
\begin{tabular}{@{}lll@{}}
\toprule
Configuration & Physics-Informed Pretraining & Supervised Training \\
\midrule
Optimizer & Adam & Adam \\
Learning rate & $3\times10^{-4}$ & $3\times10^{-4}$ \\
Epochs & 10 / 50 &  500 \\
Batch size & 256 & 256 \\
Training data & Unlabeled PDE inputs & Labeled trajectories \\
Objective & Physics-informed loss & Supervised loss \\
PI loss weights (PDE, IC, BC) & $(1,\,0.1,\,0.1)$ & -- \\
LR scheduler & -- & StepLR$(100,0.5)$ \\
Initialization & Random & Pretrained / Random \\
\bottomrule
\end{tabular}
\end{table}

At the transition to supervised fine-tuning, the pretrained parameters are retained, while the optimizer and learning-rate scheduler are reset. Only applicable physical constraints are included in the pretraining objective.

\begin{table}[pos=htbp]
\centering
\small
\setlength{\tabcolsep}{5pt}
\caption{Numbers of unlabeled pretraining inputs $N$ and labeled samples $B$ for the results in Table~\ref{tab:cross_architecture_comparison}. Supervised-only and \method{} use the same $B$ for each dataset and architecture.}
\label{tab:fno-data-counts}
\begin{tabular}{@{}lrrrrrr@{}}
\toprule
\multirow{2}{*}{Dataset} & \multicolumn{2}{c}{FNO} & \multicolumn{2}{c}{WNO} & \multicolumn{2}{c}{LNO} \\
\cmidrule(lr){2-3}\cmidrule(lr){4-5}\cmidrule(lr){6-7}
& $N$ & $B$ & $N$ & $B$ & $N$ & $B$ \\
\midrule
Burgers                & 50,000 & 10,000 & 1,000  & 1,000 & 1,000  & 1,000 \\
1D Reaction--Diffusion & 50,000 & 9,000  & 9,000  & 9,000 & 50,000 & 1,000 \\
1D Advection           & 9,000  & 9,000  & 9,000  & 9,000 & 50,000 & 1,000 \\
Diffusion--Sorption    & 9,000  & 9,000  & 9,000  & 9,000 & 50,000 & 1,000 \\
1D CFD                 & 9,000  & 9,000  & 9,000  & 9,000 & 1,000  & 1,000 \\
2D Darcy Flow          & 50,000 & 9,000  & 9,000  & 9,000 & 9,000  & 1,000 \\
AirfRANS               & 10,000 & 800    & 10,000 & 800   & 800    & 800   \\
2D Navier--Stokes      & 50,000 & 1,000  & ---    & ---   & ---    & ---   \\
\bottomrule
\end{tabular}
\end{table}




\end{document}